\documentclass[preprint,times]{elsarticle}
\usepackage[a4paper,left=1in,right=1in,top=1in,bottom=1in,footskip=.25in]{geometry}
\usepackage{subcaption}
\usepackage{amsmath}
\usepackage{amsfonts}
\usepackage{upgreek}
\usepackage{textcomp}
\usepackage{bm}
\usepackage[export]{adjustbox}

\usepackage{makecell}
\usepackage{multirow}
\usepackage[utf8]{inputenc}
\usepackage{booktabs}
\usepackage{array}
\usepackage{rotating}
\usepackage{algorithm2e}
\biboptions{sort&compress}
\usepackage{hyperref}
\usepackage{xcolor}
\usepackage{float}
\usepackage{mathrsfs}  

\begin{document}

\begin{frontmatter}

%----------------------------------------------------------------------------------------
%	TITLE SECTION
%----------------------------------------------------------------------------------------

\title{Model-Based Transfer Learning for Real-Time Damage Assessment of Bridge Networks}

%----------------------------------------------------------------------------------------
%	AUTHORS
%----------------------------------------------------------------------------------------

%\cormark[*]
%\fnmark[*]

\author[inst1]{Elisa Tomassini}
\ead{elisa.tomassini@dottorandi.unipg.it}

\author[inst2]{Enrique Garc\'{i}a-Mac\'{i}as}
\ead{enriquegm@ugr.es}

\author[inst1]{Filippo Ubertini}
\ead{filippo.ubertini@unipg.it}

\address[inst1]{Department of Civil and Environmental Engineering, University of Perugia. Via G. Duranti, 93 - 06125 Perugia, Italy.}

\address[inst2]{Department of Structural Mechanics and Hydraulic Engineering, University of Granada, Av. Fuentenueva sn, 18002 Granada, Spain.}

% ---------------------------------------------------------------
% ---------------------------------------------------------------
% ---------------------------------------------------------------
% ---------------------------------------------------------------

\begin{abstract}
\mbox{}The growing use of permanent monitoring systems has increased data availability, offering new opportunities for structural assessment but also posing scalability challenges, especially across large bridge networks. Managing multiple structures requires tracking and comparing long-term behaviour efficiently. To address this, knowledge transfer between similar structures becomes essential. This study proposes a model-based transfer learning approach using neural network surrogate models, enabling a model trained on one bridge to be adapted to another with similar characteristics. These models capture shared damage mechanisms, supporting a scalable and generalizable monitoring framework. The method was validated using real data from two bridges. The transferred model was integrated into a Bayesian inference framework for continuous damage assessment based on modal features from monitoring data. Results showed high sensitivity to damage location, severity, and extent. This approach enhances real-time monitoring and enables cross-structure knowledge transfer, promoting smart monitoring strategies and improved resilience at the network level.
\end{abstract}

% ---------------------------------------------------------------
% ---------------------------------------------------------------
% ---------------------------------------------------------------
% ---------------------------------------------------------------

% Keywords
% Each keyword is separated by \sep
\begin{keyword}
Bridge Networks \sep Damage Assessment  \sep Model-based \sep Neural Network  \sep Surrogate Models \sep Transfer Learning
\end{keyword}

\end{frontmatter}

% **************************************************************************************
% **************************************************************************************
% **************************************************************************************
% **************************************************************************************
% **************************************************************************************
% **************************************************************************************
% **************************************************************************************
% **************************************************************************************
\section{Introduction}

% General introduction
The longevity and safety of bridge infrastructure are critical to the sustainability of transportation networks and the safeguarding of public welfare. According to the 2025 report by the American Society of Civil Engineers~\cite{ReportCard2025}, there are currently over 623,000 bridges in service across the United States, with 49.1\% rated in fair condition and 6.8\% in poor condition. In China, records from 2009 to 2019 indicate that over 400 bridges experienced partial or complete collapse, with approximately 70\% of these failures ascribed to anthropogenic causes~\cite{Tan2020}. Europe faces similar challenges, with structural deficiencies reported in approximately 33\% of highway bridges~\cite{Europe2019,cuerva2021actual}. Italy, with a dense network of more than 120{,}000 bridges~\cite{Santarsiero2021}, was starkly reminded of such risks by the collapse of the Morandi Bridge in August 2018, which claimed 43 lives~\cite{calvi2019once}. These tragedies underscore the urgent need for efficient preventive maintenance and Structural Health Monitoring (SHM) systems that are scalable to entire bridge networks.

% OMA
Among various SHM methodologies, vibration-based techniques—particularly those rooted in Operational Modal Analysis (OMA)—have gained prominence due to their non-intrusive nature and suitability for in-service monitoring. By analyzing ambient excitations (e.g., traffic, environmental noise) recorded by strategically placed sensors, these methods enable the extraction of modal properties such as natural frequencies, mode shapes, and damping ratios. These properties are highly sensitive to changes in stiffness and energy dissipation, making them effective indicators of structural damage and overall integrity~\cite{Quqa2021, Cabboi2017, HE2022int}. Significant advancements in OMA algorithms have been made across time, frequency, and time-frequency domains~\cite{brincker2015introduction}, with growing emphasis on automation for continuous monitoring~\cite{rainieri2014operational}. Among the many algorithms in the literature, Covariance-driven Stochastic Subspace Identification (Cov-SSI) is one of the most widely adopted techniques~\cite{magalhaes2011explaining, rainieri2007automated}, estimating modal parameters via discrete-time state-space models visualized through stabilization diagrams. Physical modes appear as stable poles, consistently aligned across model orders, and are distinguished using hard/soft selection criteria~\cite{reynders2012fully}. Moreover, clustering algorithms~\cite{tomassini2023model, romanazzi2023iterative} further assist in grouping stable poles by frequency and mode shape similarity, enabling automated frequency tracking.

%Machine Learning Paradigms for Damage Identification
Once an effective SHM strategy is established, the subsequent critical step is the automated assessment of structural damage. Based on Rytter's classification~\cite{Rytter1993}, damage assessment comprises four fundamental tasks: detection, localization, quantification, and prognosis. These processes are essential to ensure the safety, functionality, and long-term resilience of critical infrastructure. In this field, Machine learning (ML) has emerged as a powerful tool~\cite{Farrar2012}, addressing the damage assessment tasks through three paradigms: Unsupervised Learning (UL), Supervised Learning (SL), and Semi-Supervised Learning (SSL). UL relies solely on data representing the healthy state, making it suitable when damage data is scarce or unavailable. In SL, on the other hand, models are trained using datasets labeled with both undamaged and damaged conditions. SSL represents a compromise between the two, requiring a limited set of labeled damage data to supplement the healthy-state information. Given the practical difficulties in acquiring comprehensive datasets that include damage scenarios and their ability to operate with minimal prior knowledge, UL methods have become particularly attractive for long-term real-world SHM deployments.

% Damage detection
However, UL data-driven techniques remain primarily limited to damage detection. To address this task, various statistical models have been developed to decouple environmental influences from structural changes, integrating damage detection as the first step within the statistical pattern recognition paradigm~\cite{Farrar2012}. Early work by Peeters and De Roeck~\cite{Peeters2001} used autoregressive models to analyze residuals for anomaly detection. Later, methods like Principal Component Analysis (PCA)\cite{Yan2005} and Multiple Linear Regression (MLR)\cite{Magalhaes2012, Ubertini2013, Ubertini2018a} were successfully applied to model the impact of environmental variables on modal features. Kullaa~\cite{Kullaa2003} further demonstrated the use of statistical control charts on the Z24 bridge to detect deviations by comparing predicted and measured modal parameters, providing a robust approach to distinguish environmental variability from genuine structural damage. Today, control charts are widely regarded as a state-of-the-art method for data-driven UL damage detection.

%Model Updating, Structural Identification and Surrogate modeling
SL methods offer richer insights but are hindered by the scarcity of labeled damage data. Physical testing on full-scale structures is rarely feasible, making numerical simulation the primary viable alternative. In this context, structural identification frameworks provide a complementary approach by aligning uncertain model parameters—such as stiffness constants—with experimental observations through inverse model updating~\cite{cawley2018structural}. Model updating methods are typically categorized as either deterministic or probabilistic. Deterministic approaches ~\cite{hou2021review,alkayem2018structural} involve solving an optimization problem to minimize discrepancies between measured and predicted responses through optimization procedures. While these techniques are relatively straightforward to implement and involve a low computational burden, they might be non-convex and ill-posed, thus requiring regularization techniques~\cite{titurus2008regularization} and careful definition of fitting parameters via sensitivity analyses~\cite{wan2015parameter}. Moreover, deterministic methods do not account for uncertainties, limiting their utility in risk-informed decision-making. Therefore, probabilistic approaches, such as Bayesian Model Updating (BMU) through Markov Chain Monte Carlo (MCMC)~\cite{lam2018mcmc,Li2025MCMC,Yuan2023, zhang2021model}, have been introduced to overcome this limitation. These methods naturally incorporate measurement noise, prior knowledge, and model uncertainties, effectively regularizing the inverse problem and enhancing robustness. The application of both deterministic and probabilistic system identification for SHM represents a stark challenge with large computational demands. To overcome these limitations, Surrogate Models (SM) have been introduced with the aim to provide efficient approximations of computationally expensive numerical simulations, offering high fidelity at a significantly reduced computational cost. In this framework, surrogate models act as digital twins of the structure, capable of performing continuous damage assessment~\cite{cabboi2017from, GarciaMacias2020a, garcia2019innovative, garcia2022bayesianSM,ierimonti2021transfer, GarciaMacias2025BER}.

% Starting network-scale damage assessment with damage detection through comprehensive software
Recent advancements, together with evolving regulatory frameworks, have enabled the deployment of large-scale bridge monitoring networks. In particular, integrated SHM platforms have begun to emerge, such as P3P~\cite{garcia2022p3p} and MOVA/MOSS~\cite{garcia2020mova}, developed by García-Macías \textit{et al.}, which enable automated, unsupervised, data-driven damage detection through comprehensive end-to-end workflows. A prominent example is Anas S.p.A.'s \textit{Programma SHM}\cite{ProgrammaSHM}, launched in 2022 to equip 1,000 bridges with continuous vibration-based monitoring systems by 2026, supported by an investment of €275 million. In this context, the monitoring platform was developed in collaboration with the creators of P3P~\cite{garcia2022p3p}, with the aim of enabling continuous, data-driven damage detection at a network-wide scale. While these initiatives mark a significant step forward, it is worth noting that they also introduce substantial challenges in terms of data storage, management, and computational demand. To overcome some of these obstacles, Tomassini \textit{et al.}~\cite{Tomassini2025, Tomassini2024FABRE, Tomassini2024IOMAC} proposed innovative approaches based on randomized algorithms, machine learning, and sensor grouping techniques to mitigate computational demands due to dense sensor networks. Therefore, the broad adoption of such automated, data-driven solutions within robust software ecosystems represents a crucial advancement for enabling damage detection at the scale of extensive bridge networks. Nevertheless, despite notable progress in damage detection for individual bridges, extending these methodologies to reliably perform damage localization and quantification across large and aging infrastructure stocks remains a significant challenge.

% Damage localization and quantification and PBSHM
Given the near-complete technological transfer in the domain of damage detection for networks of bridges, the current research frontier lies in advancing from detection to continuous damage localization and quantification. A major obstacle in applying ML techniques to these tasks is the need for large and diverse training datasets to ensure robust generalization across damage states. However, acquiring such data is often costly and logistically difficult. In particular, labeled data corresponding to damaged conditions are typically scarce, as real-world failures are infrequent and difficult to replicate experimentally. To address these challenges, Population-Based Structural Health Monitoring (PBSHM) has emerged in the latest years as a promising paradigm~\cite{bull2021foundations1, gosliga2021foundations2, gardner2021foundations3, gardner2022population}. PBSHM seeks to expand the effective dataset by aggregating information across a population of similar structures thus facilitating the development of generalized and transferable diagnostic tools by leveraging shared knowledge across assets. Nevertheless, differences in feature distributions and damage labels distributions across structures—arising from variations in geometry, material properties, loading conditions, or environmental influences—can undermine direct model transfer, potentially leading to significant misclassifications. To mitigate this, Transfer Learning (TL) techniques are employed~\cite{pan2010survey, Giglioni2024domainadaptation,Delo2025ifemPBSHM}. TL allows for the use of labeled data from well-characterized source domains to make reliable inferences in underrepresented or unlabeled target domains, thereby enhancing model adaptability and robustness in heterogeneous structural populations.

% What we provide to research
Building upon the previous discussion, this work introduces a novel approach based on the concept of transferable meta-models for continuous, model-driven damage identification across bridge networks. Specifically, the proposed approach employs AI-based SMs to replace high-fidelity finite element models (FEMs) with significantly improved computational efficiency. The use of neural networks as SMs offers three key advantages over traditional techniques: (i) their dynamicity in training; (ii) intrinsic capacity for TL; and (iii) minimal evaluation times, enabling the implementation of advanced damage identification algorithms such as probabilistic Bayesian methods. This allows a SM trained on one source structure to be fine-tuned and reused across similar structures within a network. By retaining knowledge about common damage mechanisms in part of the network and fine-tuning the remaining layers, the model can adapted to new target structures with a reduced training dataset, expediting the development of new SMs and facilitating network-wide damage assessment. Consequently, this methodology integrates the principles of PBSHM to propose, for the first time, a model-driven TL strategy tailored to SM-based damage identification. The proposed approach is validated on two real-world bridges, demonstrating the transferability of SMs between structures, and the ability of transferred SM to detect, localize, and quantify damage in stochastic terms across several synthetic damage scenarios. The presented results demonstrate that the proposed approach offers a scalable, more informative, and data-efficient framework capable of supporting damage inference across multiple assets. A key distinguishing feature of the proposed approach, compared to previous PBSHM studies, lies in its independence from predefined damage classification schemes. Specifically, the methodology does not require explicit labeled data for classification. Indeed, damage characterization is achieved through a regression-based formulation, which preserves the continuity of both input and output variables of the neural network, allowing for a more nuanced representation of structural conditions.

% Paper Outline
The remainder of this work is organized as follows. The theoretical background is presented in Section~\ref{section2}, followed by the detailed design of the neural network architecture and the associated TL strategy in Section~\ref{section3}. In Section~\ref{section4}, the methodology is validated through its application to both a theoretical multi-span beam case study and two real bridges: the Volumni Bridge in Italy, and the M\'endez-N\'u\~nez Bridge in Spain. Conclusions and future perspectives are discussed in Section~\ref{section5}.

\section{Theoretical background}\label{section2}

In this section, the theoretical frameworks underlying surrogate modeling and surrogate-based Bayesian damage assessment are presented.

% -------------------------------------
\subsection{Surrogate Modeling}\label{section21}

Let us define a vector of $N$ input parameters $\mathbf{x} \in \mathbb{R}^N$, where each component $x_i \in \mathbb{R}$ represents a damage-sensitive variable defined within the FEM domain. The structural response is denoted by $\mathbf{y} \in \mathbb{R}^l$, where the $l$ components may be different experimentally measurable quantities such as modal signatures, modal displacements, strains, or other quantities of interest. In the present work, particular attention is given to the construction of SMs based on the dynamic characteristics of the structure. Let us denote with $\boldsymbol{f} \in \mathbb{R}^{n}$ the vector of natural frequencies and with $\boldsymbol{\phi} \in \mathbb{R}^{m \times n}$ the corresponding mode shapes, where $n$ is the number of accounted modes and $m$ represents the number of measured degrees of freedom (DOFs i.e., the number of sensor acquisition channels deployed on the structure). In this context, $l = n \cdot (1 + m)$, accounting for a SM comprising the $n$ frequencies and the $m$ modal components of the associated mode shapes. The objective of the SM is to infer and approximate with high accuracy the functional relationship $\mathbf{y}(\mathbf{x})$, effectively capturing how variations in the structural parameters influence the structural response. \newline
The construction of a SM generally involves five main stages:

\begin{enumerate}
    \item \textbf{FEM Calibration}: A preliminary step involves the development and calibration of a high-fidelity FEM of the structure, capable of accurately reproducing its dynamic behavior. The calibration process consists of identifying and adjusting key physical and mechanical parameters that significantly influence the modal response of the system. This task is typically formulated as an optimization problem, where the objective is to minimize the discrepancy between numerical and experimental modal characteristics. For a comprehensive review of the various optimization algorithms that can be employed in this context, the reader is referred to~\cite{EREIZ2022684}.

    \item \textbf{Design Space Sampling}: The design space $\mathcal{D}$ is defined as:
\begin{equation}\label{eq:domain}
    \mathcal{D} = \left\{ \mathbf{x} \in \mathbb{R}^N : a_i \leq x_i \leq b_i ,\ i \in [1, \dots, N] \, \right\},
\end{equation}

\noindent where $\left[a_i, b_i\right]$ represent the prescribed bounds for each parameter $x_i \in \mathbf{x}$. A set of $q$ input samples $\mathbf{x}_j, j=1 \dots q$ is generated using sampling techniques such as Latin Hypercube Sampling (LHS). These samples are arranged into a design matrix $\mathbf{X} = \left[\mathbf{x}_1, \dots, \mathbf{x}_q\right]^{\rm{T}} \in \mathbb{R}^{N \times q}$. 

    \item \textbf{Generation of the training population}: Each sampled configuration $\mathbf{x}_j$ is evaluated using the calibrated FEM to determine the corresponding output $\mathbf{y}_j$. Therefore the response of the structure can be defined as $\mathbf{Y} = [\mathbf{y}_1, \dots, \mathbf{y}_q] \in \mathbb{R}^{l\times q}$, forming the training population $ \left\{ \mathbf{X}, \mathbf{Y} \right\} $. To assess the predictive performance of the SM, a separate validation set $\left\{ \mathbf{X}^v, \mathbf{Y}^v \right\}$ is sampled, where $\mathbf{X}^v \in \mathbb{R}^{N \times q^v}$ and $\mathbf{Y}^v \in \mathbb{R}^{N \times q^v}$ represent the design space and the associated response of the structure, while $q^v$ is the number of samples involved in the validation set.

    \item \textbf{Surrogate Model Construction}: The training dataset is then used to fit the surrogate model $\hat{M}$. A wide variety of models and interpolation functions can be found in the literature (see e.g.,~\cite{garcia2019innovative, garcia2022bayesianSM}).

    \item \textbf{Model Validation}: The predictions of the SM are finally compared against the FEM outputs in the validation set to evaluate its accuracy and generalization capability.
\end{enumerate}

Once validated, the SM can be embedded within an online SHM system to enable real-time tracking of damage-sensitive parameters. This is achieved by continuously mapping experimentally identified modal features onto the SM's input space.

% -------------------------------------
\subsection{Surrogate-based Bayesian damage assessment}\label{section22}

Once the SM is constructed, it is employed to continuously infer the model parameters $\mathbf{x}$, conditioned on a set of experimentally identified modal properties $\mathbf{d}(t) \in \mathbb{R}^{l}$ at each time instant $t$. The vector $\mathbf{d}(t)$ includes the identified natural frequencies $f_r(t)$ for $r = 1, \dots, n$ and the corresponding mode shapes $\boldsymbol{\phi}_r(t) \in \mathbb{R}^m$. These quantities are periodically updated over time $t$ based on monitoring data.

Bayesian inference is adopted to estimate the posterior distribution of the model parameters $\mathbf{x}(t)$, conditioned on the surrogate model $\hat{M}$ and the identified modal data $\mathbf{d}(t)$. According to Bayes’ theorem:
\begin{equation}
p(\mathbf{x}(t)|\mathbf{d}(t), \hat{M}) = \frac{p(\mathbf{d}(t)|\mathbf{x}(t), \hat{M}) \, p(\mathbf{x}(t)|\hat{M})}{p(\mathbf{d}(t)|\hat{M})},
\end{equation}
where: $p(\mathbf{d}(t)|\mathbf{x}(t), \hat{M})$ is the likelihood function, $p(\mathbf{x}(t)|\hat{M})$ is the prior distribution and $p(\mathbf{d}(t)|\hat{M})$ is the evidence (or marginal likelihood) at time instant $t$.

In this work, the prior distribution $p(\mathbf{d}(t)|\hat{M})$ at each time instant $t$ is assumed to be a truncated Gaussian distribution with mean equal to the nominal (undamaged) values of the fitting parameters and boundaries defined according to Eq.~(\ref{eq:domain}).

\begin{algorithm}[H]
\caption{MCMC for Surrogate-based Bayesian Inference} \label{MCMCalgorithm}
\KwIn{Initial fitting parameters $\mathbf{x}_0$, initial covariance matrix $\Sigma_0$, standard deviations of the frequencies $\boldsymbol{\sigma}_f$, standard deviations of the mode shapes $\boldsymbol{\sigma}_\phi$, scaling factor $s_N = 2.38^2/N$, number of samples $N_s$, burn-in $N_b$.}
\KwOut{Markov chain samples $\{ \mathbf{x}_j \}$ and posterior statistics $\left\{ \hat{\boldsymbol{\mu}}, \hat{\boldsymbol{\sigma}} \right\}$.}

Set $\mathbf{x}_c = \mathbf{x}_0$, $\Sigma_p = \Sigma_0$\;

Evaluate posterior $p(\mathbf{x}_c | \mathbf{d}(t), \hat{M})$ for the starting point\;

Initialize matrix $\hat{\mathbf{x}}$ as empty\;

\For{$i = 1$ \KwTo $N_s$}{
    Propose sample $\mathbf{x}_p \sim \mathcal{N}(\mathbf{x}_c, \Sigma_p)$\ 

    Evaluate posterior $p(\mathbf{x}_p | \mathbf{d}(t), \hat{M})$\;

    Compute acceptance probability:\;
    $\alpha = \min\left(1, \frac{p(\mathbf{x}_p | \mathbf{d})}{p(\mathbf{x}_c | \mathbf{d})} \right)$\;

    Sample $u \sim \mathcal{U}(0,1)$\;
    
    \If{$u < \alpha$}{
        Set $\mathbf{x}_c = \mathbf{x}_p$\;
    }

    \If{$i > N_b$}{
        Update proposal covariance:\;
        $\Sigma_p = s_N \cdot \mathrm{cov}(\mathbf{x}_1, ..., \mathbf{x}_i)$\;

        Append $\mathbf{x}_p$ to $\hat{\mathbf{x}}$\;
    }
}

Compute posterior statistics:\;
$\hat{\boldsymbol{\mu}} = [\mu_1, \dots, \mu_N]$, $\hat{\boldsymbol{\sigma}} = [\sigma_1, \dots, \sigma_N]$\;

With: $\mu_j = \mathbb{E}[\hat{\mathbf{x}}_j], \quad \sigma_j = \sqrt{\mathbb{E}[\hat{\mathbf{x}}_j^2] - \mu_j^2}, \quad j = 1, \dots, N$\;

\end{algorithm}

The likelihood quantifies the probability of observing the measured modal data $\mathbf{d}(t)$ given a set of model parameters $\mathbf{x}(t)$ by mean of the surrogate model $\hat{M}$. Assuming normal Gaussian distributions for frequency estimates, the relationship for the $r$-th mode is:
\begin{equation}
f_r = \hat{f}_r(\mathbf{x}) + \varepsilon_{f_r}, \quad \varepsilon_{f_r} \sim \mathcal{N}(0, \sigma^2_{f_r}),
\end{equation}
where $\hat{f}_r(\mathbf{x})$ is the predicted frequency from the surrogate model $\hat{M}$. The corresponding likelihood is:
\begin{equation}\label{likelihood_frequencies}
p(f_r | \mathbf{x}) = \frac{1}{\sqrt{2\pi \sigma^2_{f_r}}} \exp\left(-\frac{(f_r - \hat{f}_r(\mathbf{x}))^2}{2\sigma^2_{f_r}}\right),
\end{equation}
where $\sigma_{f_r}$ is the standard deviation of the $r$-th frequency.

For the mode shapes, assuming uncorrelated Gaussian errors and introducing a scaling factor $\beta_r(\mathbf{x})$ between the experimental and predicted mode shapes:
\begin{equation}
\boldsymbol{\phi}_r = \beta_r(\mathbf{x}) \boldsymbol{\hat{\phi}}_r(\mathbf{x}) + \varepsilon_{\boldsymbol{\phi}_r}, \quad \varepsilon_{\boldsymbol{\phi}_r} \sim \mathcal{N}(0, \sigma^2_{\boldsymbol{\phi}_r} \Sigma_{\boldsymbol{\phi}_r}), \quad \textrm{with: \;}\beta_r(\mathbf{x}) = \frac{\boldsymbol{\phi}_r^\textrm{T}  \boldsymbol{\hat{\phi}}_r(\mathbf{x})}{\boldsymbol{\hat{\phi}}_r(\mathbf{x})^\textrm{T}  \boldsymbol{\hat{\phi}}_r(\mathbf{x})}, \quad \Sigma_{\boldsymbol{\phi}_r} = \frac{\boldsymbol{\phi}_r^\textrm{T}  \boldsymbol{\phi}_r}{m} \mathbf{I}_m.
\end{equation}
Then, the likelihood of $\boldsymbol{\phi}_r$ becomes:
\begin{equation}
p(\boldsymbol{\phi}_r | \mathbf{x}) = \frac{1}{\sqrt{(2\pi)^m \det \Sigma_{\boldsymbol{\phi}_r}}} \exp\left(-\frac{1}{2} (\boldsymbol{\phi}_r - \beta_r(\mathbf{x}) \boldsymbol{\hat{\phi}}_r(\mathbf{x}))^\textrm{T} \Sigma^{-1}_{\boldsymbol{\phi}_r} (\boldsymbol{\phi}_r - \beta_r(\mathbf{x}) \boldsymbol{\hat{\phi}}_r(\mathbf{x})) \right).
\end{equation}
Alternatively, using a MAC-based formulation the likelihood can be expressed as:
\begin{equation}\label{likelihood_MAC}
p(\boldsymbol{\phi}_r | \mathbf{x}) = \frac{1}{\sqrt{2\pi \sigma^2_{\boldsymbol{\phi}_r}}} \exp\left(-\frac{1 - \text{MAC}_r}{2\sigma^2_{\boldsymbol{\phi}_r}}\right),
\end{equation}
where $\sigma_{\boldsymbol{\phi}_r}$ stands for the standard deviation associated to the the $r$-th mode shape.

The total likelihood over all $n$ modes is:
\begin{equation}\label{likelihood_total}
p(\mathbf{d}|\mathbf{x}) = \prod_{r=1}^{n} p(f_r | \mathbf{x}) \cdot p(\boldsymbol{\phi}_r | \mathbf{x}).
\end{equation}

The posterior distribution $p(\mathbf{x}(t)|\mathbf{d}(t), \hat{M})$ is sampled using an MCMC approach. The implementation (outlined in Algorithm~\ref{MCMCalgorithm}) requires the prior specification of the initial fitting parameter vector $\mathbf{x}_0$ and covariance matrix $\Sigma_0$, the standard deviations $\boldsymbol{\sigma}_f$ and $\boldsymbol{\sigma}_\phi$ of frequencies and mode shapes to be used in the likelihood estimation and the total number of samples $N_s$ to be drawn by the MCMC. Once the algorithm starts, it is needed a stabilization period is necessary where it can explore the space of $\mathbf{x}$, denoted as burn-in period. After the burn-in period of length $N_b$, the Markov chain is supposed to be stable and so the covariance matrix of the proposal fitting parameters can be adapted considering the covariances of the previous samples scaled by the factor $s_N = 2.38^2/N$ as suggested by Haario \textit{et al.}~\cite{HaarioDRAM}. This Bayesian inference process is applied independently at each time instant $t$ when new modal data $\mathbf{d}(t)$ become available, in order to estimate and track the evolution of the posterior statistics (for instance, the mean $\hat{\boldsymbol{\mu}}(t)$ and the standard deviation $\hat{\boldsymbol{\sigma}}(t)$ of the fitting parameters $\mathbf{x}$) over time.

%%%%%%%%%%%%%%%%%%%%%%%%%%%%%%%%%%%%%%%%%%%%%

\section{Creation of surrogate models via Transfer Learning}\label{section3}
\begin{figure}[H]
\centering
   \includegraphics[width=1\textwidth]{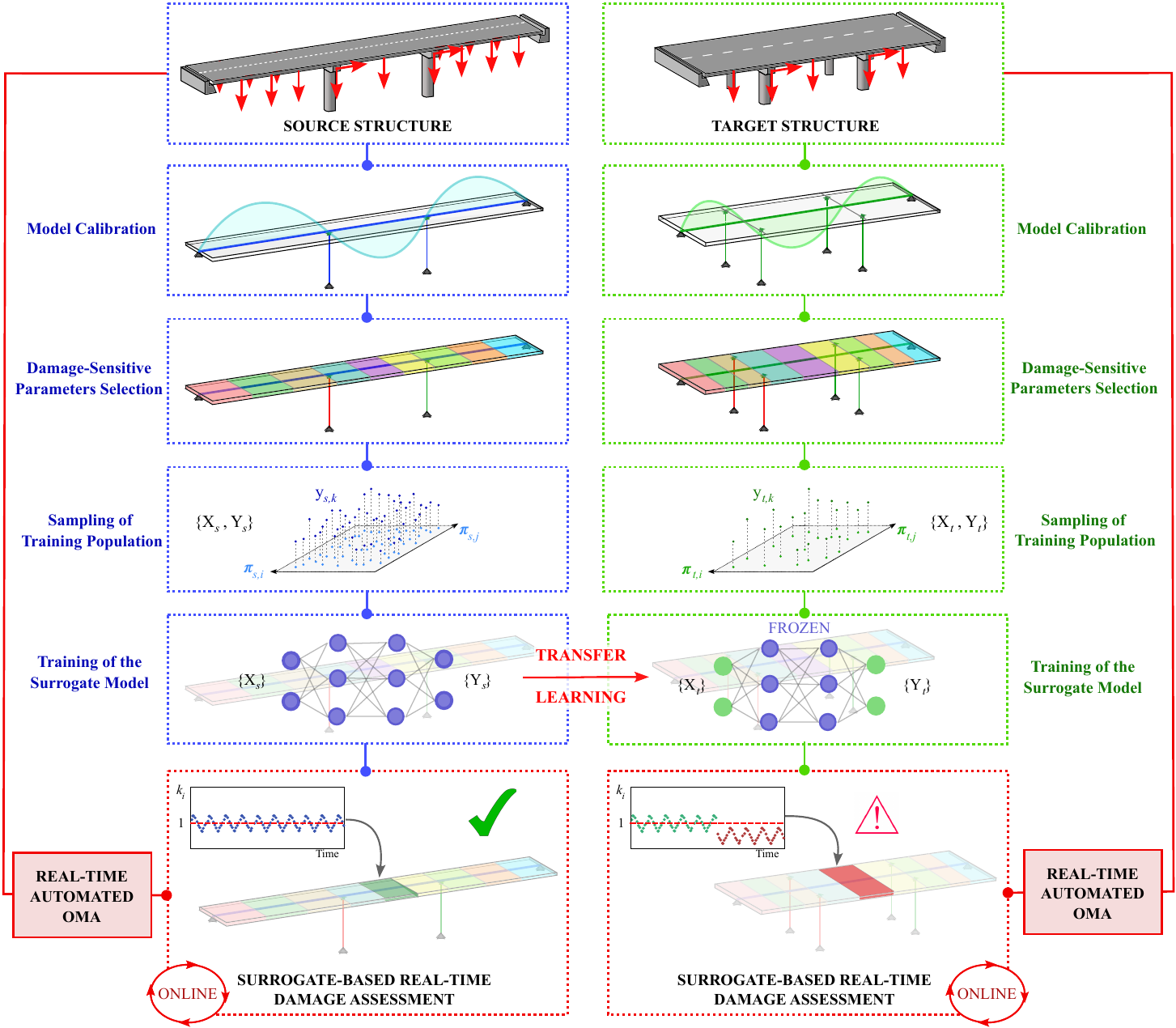}
   \caption{Flowchart of the proposed model-based transfer learning methodology.}    
    \label{fig:SKETCH_METHODOLOGY}
\end{figure}
The overall flowchart of the proposed methodology is shown in Fig.~\ref{fig:SKETCH_METHODOLOGY}. The core idea is to adapt and reuse a SM trained on a source structure to generate a SM of a similar (target) structure within a network. To this end, a SM is first constructed for the source structure using a Deep Feedforward Neural Network (FNN) trained on a rich database generated by forward FEM evaluations in order to accurately predict its modal properties. Afterwards, a considerably reduced training database is created for the target structure, on which the previously constructed source SM is adapted through fine-tuning. In this way, it is possible to transfer the knowledge acquired from the source structure, requiring a minimum number of training samples for the target structure and thus considerably accelerating the entire process.

The use of FNNs to construct SMs is pivotal in this application, as their architecture inherently facilitates knowledge transfer across multiple structural assets. While numerous non-intrusive surrogate modeling techniques are available in the literature (as discussed in Section~\ref{section21}), FNN-based models offer specific advantages. On one hand, FNNs are capable of capturing complex relationships between damage-sensitive parameters and structural behavior, as further elaborated in Section~\ref{section31}. On the other hand, due to their internal structure, FNNs can be efficiently adapted to similar structural systems through fine tuning transfer learning. This is achieved by freezing selected inner layers of a pretrained network on a source structure, thus preserving generalized knowledge of typical damage mechanisms, while fine tuning the remaining specialized layers to learn the distinct characteristics of a target structure with typological similarities (see Section~\ref{section32}). This strategy enables effective model reuse and supports scalable SHM across large infrastructure networks.

% -------------------------------------
\subsection{Neural Network architecture}\label{section31}

\begin{figure}[H]
\centering
   \includegraphics[width=1\textwidth]{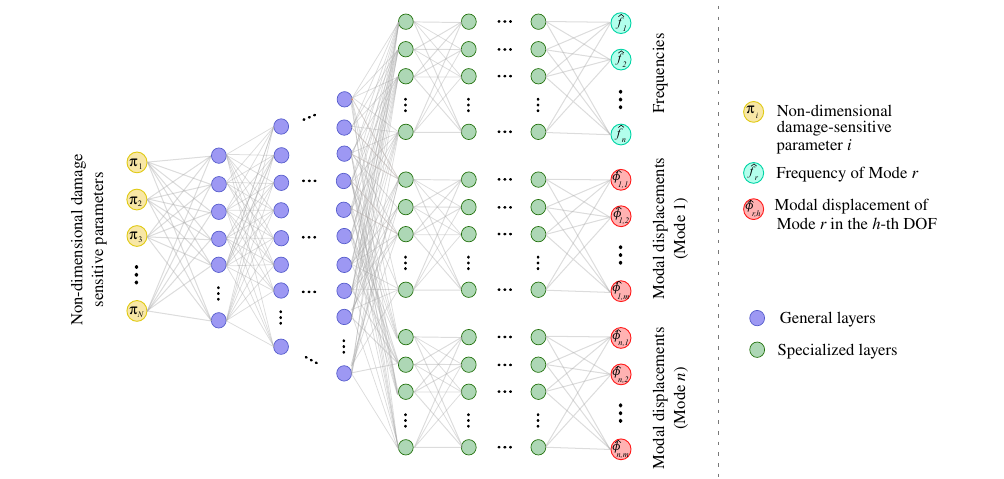}
   \caption{Neural Nework architecture.}    
    \label{fig:NN_model}
\end{figure}

The architecture of the feedforward neural network (FNN), illustrated in Figure~\ref{fig:NN_model}, adopts a modular configuration that enables the simultaneous prediction of both natural frequencies and mode shapes, accounting for the distinct characteristics of each modal component.

In this work, it is assumed that structural damage can be represented by variations in the elastic moduli $E_i, i = 1, \dots , N$, across $N$ specific regions of the structure. Accordingly, the damage is described by a set of stiffness multipliers $k_i$, each one associated with the elastic modulus $E_i$ of the $i$-th region.

The input to the network is a vector of non-dimensional damage-sensitive parameters $\boldsymbol{\pi} \in \mathbb{R}^N$, associated with the $N$ selected design variables $E_i$. The components of $\boldsymbol{\pi}$ are defined as:
\begin{equation}\label{eq:pi}
\pi_i = \frac{1}{\bar{\bar{f_1}}L^2} \sqrt{\frac{k_i E_i I}{\rho A}}, \quad i = 1, \dots, N,
\end{equation}
where $\bar{\bar{f_1}}$ is the fundamental frequency from the FEM, $L$ is the average span length of the bridge, $k_i$ is the stiffness multiplier representing damage in the elastic modulus $E_i$, $\rho$ is the average mass density of the bridge deck, and $A$ and $I$ denote the average cross-sectional area and moment of inertia of the bridge deck, respectively. Thus, referring to Eq.~(\ref{eq:domain}), we set $\mathbf{x} \equiv \boldsymbol{\pi}$, and define the sampling space as $\mathbf{X} = \left[ \boldsymbol{\pi}_1, \dots, \boldsymbol{\pi}_q \right] \in \mathbb{R}^{N \times q}$.

The FNN output provides the modal signatures of the structure, including both the natural frequencies and the corresponding mode shapes, with a dedicated neuron assigned to each frequency and real modal displacement within each mode shape. Specifically, $\mathbf{Y}$ contains the frequencies and mode shapes obtained from FEM simulations to construct the training population, and is structured as:
\begin{equation}
\mathbf{Y} = 
\begin{bmatrix}
\mathbf{y}_1  & \mathbf{y}_2  & \cdots & \mathbf{y}_q
\end{bmatrix}
=
\begin{bmatrix}
\boldsymbol{\overline{f}}_1 & \boldsymbol{\overline{f}}_2 & \cdots & \boldsymbol{\overline{f}}_q \\
\boldsymbol{\overline{\phi}}_{1,1} & \boldsymbol{\overline{\phi}}_{1,2} & \cdots & \boldsymbol{\overline{\phi}}_{1,q} \\
\vdots & \vdots & \ddots & \vdots \\
\boldsymbol{\overline{\phi}}_{n,1} & \boldsymbol{\overline{\phi}}_{n,2} & \cdots & \boldsymbol{\overline{\phi}}_{n,q}
\end{bmatrix}
\in \mathbb{R}^{l \times q},
\end{equation}

\noindent where $\boldsymbol{\overline{f}}_k = \left[ \overline{f}_1, \overline{f}_2, \dots, \overline{f}_n \right]^{\rm{T}}_k$ and $\boldsymbol{\overline{\phi}}_{r,k} = \left[ \overline{\phi}_{r,1}, \overline{\phi}_{r,2}, \dots, \overline{\phi}_{r,m} \right]^{\rm{T}}_k$ represent the frequency vector and the $r$-th mode shape for $r = 1, \dots , n$ of the $k$-th sample associated to the input $\boldsymbol{\pi}_k$ for $k = 1, \dots , q$, respectively.

The network begins with an input layer that receives the $N$ non-dimensional stiffness parameters $\boldsymbol{\pi}$, followed by multiple fully connected (dense) layers using hyperbolic tangent activation functions. These layers, referred to as \textit{general layers}, are responsible for capturing the nonlinear relationships between the input damage-sensitive parameters and the modal responses. The architecture then branches into two distinct modules, defined as \textit{specialized layers}:
\begin{itemize}
\item The first branch predicts the natural frequencies $\hat{f}_r$, for $r = 1, \dots, n$, using a sequence of layers with Gaussian Error Linear Unit (GELU) activation functions, promoting smooth and stable learning of positive-valued outputs.
\item The second branch consists of $n$ independent subnetworks, each dedicated to predicting the $r$-th mode shape. Each subnetwork outputs a vector of $m$ modal displacements and uses hyperbolic tangent activation functions to capture the spatial variation and distribution of the respective mode shape.
\end{itemize}

This modular architecture allows the network to specialize its learning process, effectively addressing the physical diversity and spatial complexity of individual modal components. As a result, the surrogate model $\hat{M}$ accurately reproduces the dynamic response of the structure—both frequencies and mode shapes—based solely on changes in stiffness distribution as described by the non-dimensional damage parameters $\boldsymbol{\pi}$.

Finally, the optimization problem is formulated as:
\begin{equation}
\hat{\boldsymbol{\pi}} = \arg\min_{\boldsymbol{\pi}} \; \mathcal{L}(\boldsymbol{\pi}),
\end{equation}
where the loss function $\mathcal{L}(\boldsymbol{\pi})$ is composed of contributions from both frequencies and mode shapes:
\begin{equation}\label{eq.loss}
\mathcal{L}(\boldsymbol{\pi}) = \mathcal{L}_f (\boldsymbol{\pi})+ \mathcal{L}_\phi (\boldsymbol{\pi}),
\end{equation}
with $\mathcal{L}_f (\boldsymbol{\pi})$ and $\mathcal{L}_\phi (\boldsymbol{\pi})$ defined as:
\begin{equation}
\begin{split}
\mathcal{L}_f (\boldsymbol{\pi}) &= {\sum_{r=1}^{n} c_r \left| 1 - \beta^{\, |\hat{f}_r(\boldsymbol{\pi}) - \overline{f}_r|} \right|}, \\
\mathcal{L}_\phi (\boldsymbol{\pi}) &= \sum_{r=1}^{n} d_r \left[ 1 - \mathrm{MAC} \left( \hat{\boldsymbol{\phi}}_r(\boldsymbol{\pi}), \boldsymbol{\overline{\phi}}_r \right) \right],
\end{split}
\end{equation}
where $\overline{f}_r$ and $\boldsymbol{\overline{\phi}}_r$ denote the $r$-th frequency and mode shape from $\mathbf{Y}$, respectively. The parameter $\beta > 1$ is a scaling factor for $\mathcal{L}_f$, which can be tuned alongside the weighting coefficients $c_r$ and $d_r$.

% -------------------------------------
\subsection{Transfer learning}\label{section32}

In this study, knowledge transfer is performed from a SM developed for a \textit{source structure} to a \textit{target structure}, for which no SM is available. Specifically, the objective is to leverage the knowledge embedded in the source surrogate model to construct a new one for the target structure. The main advantage of this approach lies in its efficiency: generating a sufficiently rich dataset for training a SM can be computationally expensive—especially for complex structures characterized by nonlinear behavior, interface reactions, or highly refined discretizations. In such cases, the computational burden of building a dedicated training dataset - or worse, the direct FEM updating - becomes significant. By starting from a pre-trained SM developed for a structurally similar source system, the size of the dataset required for the target system can be substantially reduced, minimizing both time and computational costs.

Let us define two domains, according to Eqs.~(\ref{eq:domain}) and (\ref{eq:pi}), as:
\begin{equation}
\begin{split}
    \mathcal{D}_s &= \left\{ \boldsymbol{\pi}_s \in \mathbb{R}^N : a_{s,i} \leq \pi_{s,i} \leq b_{s,i},\ i \in [1, \dots, N] \right\}, \\
    \mathcal{D}_t &= \left\{ \boldsymbol{\pi}_t \in \mathbb{R}^N : a_{t,i} \leq \pi_{t,i} \leq b_{t,i},\ i \in [1, \dots, N] \right\},
\end{split}
\end{equation}
where the subscripts \textit{s} and \textit{t} refer to the source and target structures, respectively. The corresponding design matrices are then:
\begin{equation}
\begin{split}
\mathbf{X}_s &= \left[ \boldsymbol{\pi}_{s,1}, \dots, \boldsymbol{\pi}_{s,q_s} \right] \in \mathbb{R}^{N \times q_s}, \\
\mathbf{X}_t &= \left[ \boldsymbol{\pi}_{t,1}, \dots, \boldsymbol{\pi}_{t,q_t} \right] \in \mathbb{R}^{N \times q_t},
\end{split}
\end{equation}
where $q_s$ and $q_t$ represent the size of the sampling spaces generated via LHS for the source and target structures, respectively.

Considering the FEM of both structures, the corresponding training populations can be defined as $\left\{ \mathbf{X}_s, \mathbf{Y}_s \right\} $ and $\left\{ \mathbf{X}_t, \mathbf{Y}_t \right\}$, where $\mathbf{Y}_s \in \mathbb{R}^{l_s \times q_s}$ and $\mathbf{Y}_t \in \mathbb{R}^{l_t \times q_t}$ represent the outputs of the source and target FEMs, corresponding to the input sets $\mathbf{X}_s$ and $\mathbf{X}_t$, respectively.

The neural network proposed in Section~\ref{section31} is first used to develop the surrogate model $\hat{M}_s$ based on the source training population $\left\{ \mathbf{X}_s, \mathbf{Y}_s \right\}$. TL is then performed through fine-tuning. During this phase, the general layers of the network used to produce $\hat{M}_s$ are frozen, and only the specialized layers are re-trained using the target training population $\left\{ \mathbf{X}_t, \mathbf{Y}_t \right\}$. This allows the model to focus on learning target-specific features while retaining the general knowledge acquired from the source domain. For TL to be effective and to prevent overfitting, it is required that $q_s > q_t$. This condition also enables the rapid construction of the SM of the target structure using a limited number of evaluations, thereby significantly reducing the computational burden associated with the generation of extensive training population and SM training, which are typically the most computationally-intensive phases. 

For TL to be viable, it is essential to establish clear criteria for structural similarity and consistent parameter definitions, ensuring that the SM can generalize from the source structure to the target. In this work, two key similarity criteria are considered:

\begin{enumerate}
    \item \textbf{Geometrical similarity:} This refers to the overall structural resemblance between the two systems. In the present case, this primarily concerns the number of spans. Differences in the dimensions of structural elements are handled through the use of the non-dimensional damage-sensitive parameters defined in Section~\ref{section31}, which allow the model to dynamically adapt to variations in properties such as span lengths, elastic moduli, and mass distributions.
    
    \item \textbf{Consistency in damage parameter distribution:} The number of non-dimensional damage parameters $N$ in input to the network must be identical for both the source and target structures. Additionally, the relative positions and the relative extent of the controlled regions must be preserved. This requirement is particularly crucial for the transfer of mode shapes.
\end{enumerate}

The architecture introduced in Section~\ref{section31} is highly generalizable across structures that satisfy these similarity criteria. This enables the development of complex networks trained on a large number of structurally similar cases, thereby enhancing the model’s ability to learn and generalize damage mechanisms across diverse structural scenarios.

Additionally, the modular nature of the network’s output allows for the selective activation of output branches. This feature is especially useful when the target FEM includes a different number of modes compared to the source model. In such cases, the branches corresponding to unused or unmatched modes can be deactivated, and the frequency output vector can be resized accordingly.

Regarding differences in sensor placement—which affect the size and structure of the mode shape vectors—similarity can be preserved by constructing the TL space using the full set of modal displacements defined in the source model, not just those corresponding to shared sensor locations. This broader perspective enhances TL by leveraging a richer set of modal information. However, for continuous damage assessment, it is still necessary to use the actual sensor configuration of the target structure.

%%%%%%%%%%%%%%%%%%%%%%%%%%%%%%%%%%%%%%%%%%%%%

\section{Numerical results and discussion}\label{section4}

% -------------------------------------
\subsection{Numerical case study: 5-spans girder}\label{section41}

\begin{figure}[H]
\centering
   \includegraphics[scale=1]{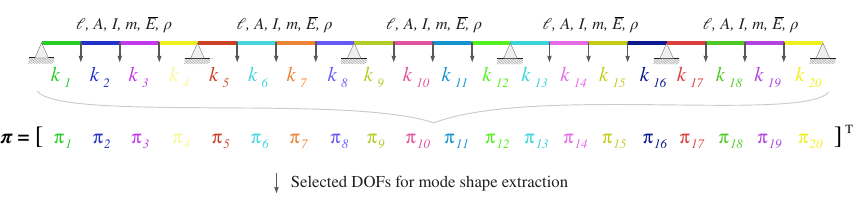}
   \caption{Theoretical multi-span beam model.}    
    \label{fig:Beam_model}
\end{figure}

To illustrate the approach and validate the performance of the proposed NN in a simplified context, the methodology described in Section~\ref{section3} is first applied to a case study involving two five-span continuous beams modeled in SAP2000. The static configuration of the systems is depicted in Figure~\ref{fig:Beam_model}.
Both multi-span beam models are defined by span lengths equal to $\mathscr{l}$ and are characterized by rectangular cross-sections. This configuration ensures that the average values of the key geometrical parameters—specifically, the span length $L$, cross-sectional area $A$, moment of inertia $I$—are equal to their corresponding local values. Same assumption is adopted for the mass density $\rho$. Each span is subdivided into four equal-length control regions, each having a length of $\mathscr{l}/4$ and associated with an individual stiffness multiplier $k_i$, for $i = 1, \dots, N$, resulting in a total of $N = 20$ design parameters uniformly distributed along the entire beam. A uniform reference Young’s modulus $\overline{E}$ is assumed for the undamaged structure. Consequently, under the assumption of constant geometric and mass properties, damage can be modeled as a reduction in stiffness, described by the relation $E_i = k_i \overline{E}$ for $i = 1, \dots, N$. Note that the reference (undamaged) configuration is defined accordingly by setting $k_i = 1$ for all $i = 1, \dots, N$. The relevant properties and parameter configurations for this test case are summarized in Table~\ref{tab:5spans_beam}.

\begin{table}[b]
\setlength{\tabcolsep}{3pt} %% default is 6pt					
\newcommand\Tstrut{\rule{0pt}{0.3cm}}         % = `top' strut
\newcommand\Bstrut{\rule[-0.15cm]{0pt}{0pt}}   % = `bottom' strut	
 \footnotesize		
 \caption{Key geometric and physical characteristics of the source and target beam models.}
 \vspace{0.1cm}
   \centering
   \begin{tabular}{m{5cm}
   >{\centering\arraybackslash}m{0.8cm}
   >{\centering\arraybackslash}m{3cm}
   >{\centering\arraybackslash}m{3cm}
   }
   \hline
    \multicolumn{2}{c}{Parameter} & Source & Target \Tstrut\Bstrut\\
   \hline
   Span length & $\mathscr{l}$ & 8 m & 5 m \Tstrut\\
   Average span length & $L$ & 8 m & 5 m \Tstrut\\
   Average cross - sectional Area & $A$ & 300x500 mm & 500x500 mm \Tstrut\\
   Average Inertia & $I$ & 3.125E-3 & 5.208E-3 \Tstrut\\
   Reference Elastic Modulus & $\overline{E}$ & 35000 MPa & 32000 MPa \Tstrut\\
   Mass per unit volume & $\rho$ & 2.55 ton/m$^3$ & 2.55 ton/m$^3$ \Tstrut\\
   Fundamental frequency of the FEM & $\overline{f}$ & 13.06 Hz & 25.52 Hz \Tstrut\\
   Boundaries of the damage multipliers & $k_i$ & $[0.70, 1.05]$ & $[0.70, 1.05]$ \Tstrut\\
   \hline
   \end{tabular}
   \label{tab:5spans_beam}
\end{table}

\begin{figure}[t]
\centering
   \includegraphics[width=1\textwidth]{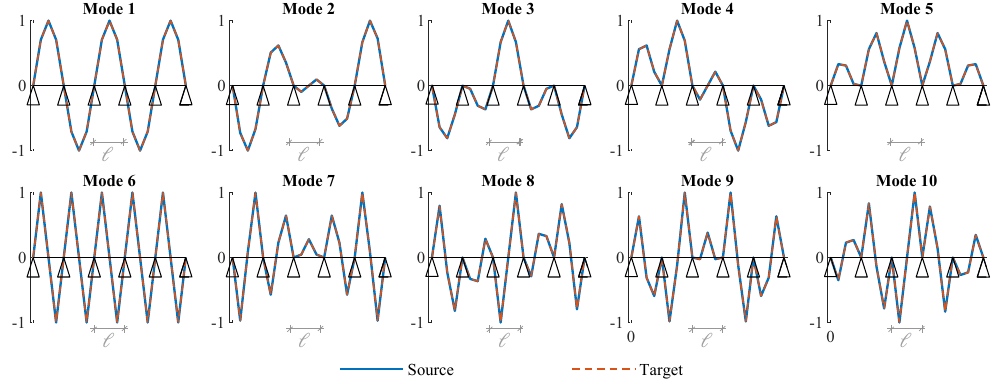}
\caption{Mode shapes of the multi-span beam models: representative modal displacements sampled at controlled locations along the spans for both source and target structures.}
    \label{fig:modes_beams}
\end{figure}

\begin{table}[t]
\setlength{\tabcolsep}{3pt} %% default is 6pt					
\newcommand\Tstrut{\rule{0pt}{0.3cm}}         % = `top' strut
\newcommand\Bstrut{\rule[-0.15cm]{0pt}{0pt}}   % = `bottom' strut	
 \footnotesize		
\caption{Reference natural frequencies for the source and target beam models in the undamaged state.}
 \vspace{0.1cm}
   \centering
   \begin{tabular}{>{\centering\arraybackslash}m{1.5cm} 
    >{\centering\arraybackslash}m{2cm} 
    >{\centering\arraybackslash}m{2cm} 
    >{\centering\arraybackslash}m{0.1cm} 
    >{\centering\arraybackslash}m{1.5cm} 
    >{\centering\arraybackslash}m{2cm} 
    >{\centering\arraybackslash}m{2cm} 
}
   \hline
    & \multicolumn{2}{c}{Frequency [Hz]} & & & \multicolumn{2}{c}{Frequency [Hz]} \Tstrut\Bstrut\\
 \cmidrule{2-3} \cmidrule{6-7}
   Mode & Source & Target & & Mode & Source & Target \Tstrut\Bstrut\\
   \hline
    1 &   13.06 & 25.52 & & 6 & 51.00 & 99.85 \Tstrut\\
    2 &   14.46 & 28.22 & & 7 & 53.58 & 104.94 \Tstrut\\
    3 &   18.00 & 35.07 & & 8 & 59.44 &  116.67 \Tstrut\\
    4 &   22.55 & 43.85 & & 9 & 66.19 & 130.56 \Tstrut\\
    5 &   26.91 & 52.24 & & 10 & 72.05 & 143.05 \Tstrut\\
       \hline
   \end{tabular}
   \label{tab:beam_frequencies}
\end{table}

An LHS strategy was adopted to generate the design spaces for both the source and the target structures. The LHS procedure was applied by sampling the stiffness multipliers $k_i$ within the bounds $[ a_i, b_i]$ reported in Table~\ref{tab:5spans_beam}. For each sampled configuration, the corresponding non-dimensional damage-sensitive parameters $\boldsymbol{\pi}$ were computed using Eq.~(\ref{eq:pi}), thereby constructing the input design spaces $\mathbf{X}_s$ and $\mathbf{X}_t$. Specifically, the design matrix for the source structure $\mathbf{X}_s$, consists of $q_s = 2048$ samples, while the design matrix for the target structure $\mathbf{X}_t$ consists of $q_t = 512$ samples. The associated modal properties of the source $\mathbf{Y}_s$ and of the target $ \mathbf{Y}_t$ structures were subsequently obtained through Monte Carlo simulations performed on the FEMs of the respective structures.

For both the source and target domains, the first $n = 10$ mode shapes of the two beam models were employed to validate the proposed methodology. Table~\ref{tab:beam_frequencies} reports the reference natural frequencies—corresponding to the undamaged configuration—for both the source and target structures, while Figure~\ref{fig:modes_beams} depicts the corresponding mode shapes. Each mode shape was sampled by extracting three modal displacements per span, located at positions $\mathscr{l}/4$, $\mathscr{l}/2$, and $3\mathscr{l}/4$, as defined in Figure~\ref{fig:Beam_model}, resulting in a total of $m = 15$ modal displacements per mode shape. Thus, the structural response matrices $\mathbf{Y}_s$ and $\mathbf{Y}_t$—associated respectively with the source and target structures—are constructed from a total of $l = 160$ output parameters: 10 natural frequencies and 150 modal displacements (i.e., 15 components for each of the 10 selected modes).

\begin{figure}[H]
\centering
   \includegraphics[width = 1\textwidth]{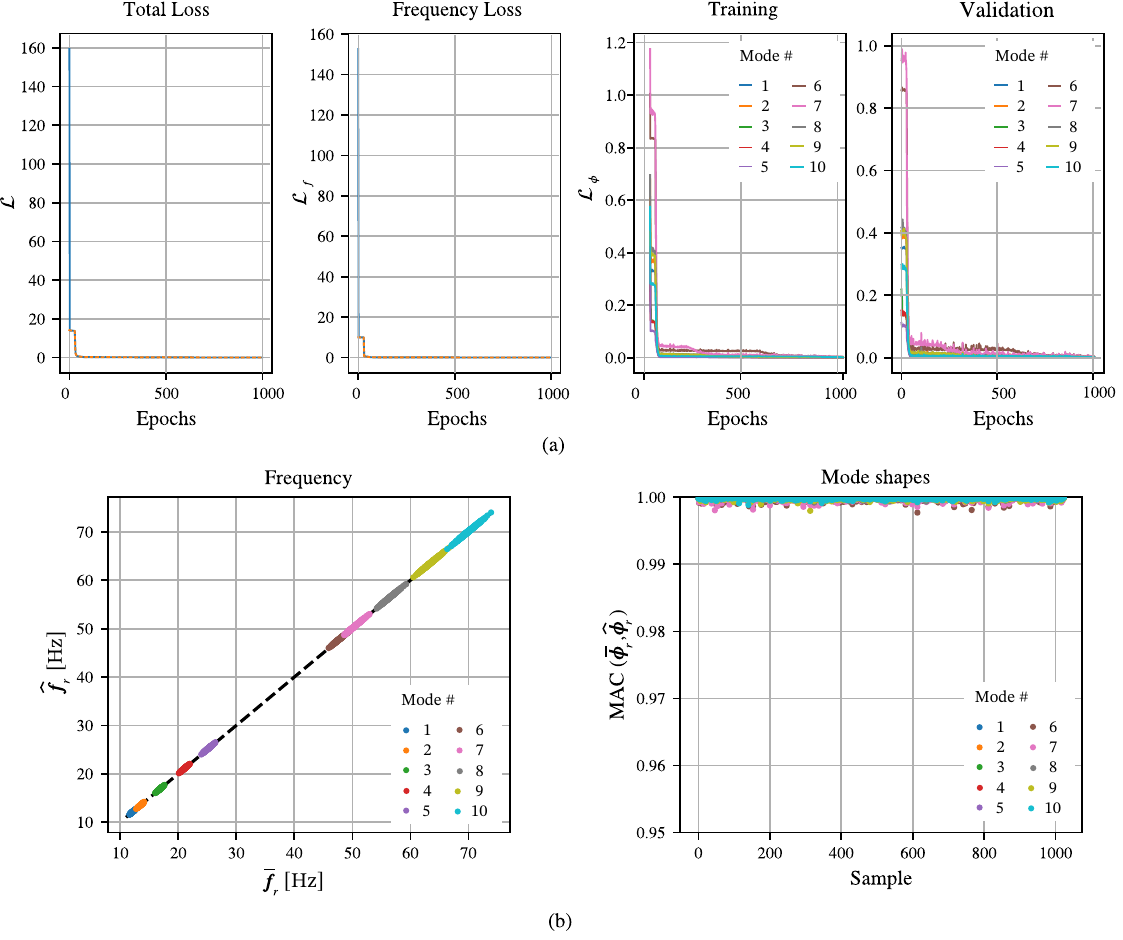}
\caption{Training of the surrogate model $\hat{M}_s$ developed for the source beam model. (a) Total loss and individual losses associated with frequency and mode shape predictions during the training phase. (b) Comparison between the predicted frequencies $\hat{f}$ and the FEM reference values $\overline{f}$, along with the MAC values computed between the predicted mode shapes $\hat{\phi}$ and the FEM mode shapes $\overline{\phi}$ for each sample of the test set.}
    \label{fig:training_beams}
\end{figure}

The FNN proposed in Section \ref{section31} was trained for a total of 1000 epochs on the source training dataset $\left\{ \mathbf{X}_s, \mathbf{Y}_s \right\}$, with 20\% of the data randomly selected and held out for validation purposes. The training process employed the custom loss function defined in Eq.~(\ref{eq.loss}), where the weighting coefficients $c_r$ and $d_r$, for $r = 1, \dots, n$, were set to unity, and the scaling factor $\beta$ was assigned a value of 100. The architecture's hyperparameters were optimized through a series of parametric analyses aimed at balancing convergence behavior on both training and validation datasets with the overall prediction accuracy of the natural frequencies and corresponding mode shapes. The network was trained using the Adam optimization algorithm, configured with an initial learning rate of $10^{-3}$ and a momentum coefficient of 0.98. A mini-batch size of 16 was adopted, with the dataset being shuffled at the beginning of each epoch to enhance generalization and avoid training bias. The total training and validation losses, along with the individual contributions from the frequency and mode shape components, are reported in Figure~\ref{fig:training_beams}(a).

To assess the generalization capability of the FNN trained on the source structure, a validation dataset $\left\{ \mathbf{X}_s^v, \mathbf{Y}_s^v \right\}$ was generated using the same LHS strategy adopted for the training dataset, consisting of $q_s^v = 512$ samples. Figure~\ref{fig:training_beams}(b) also presents the validation results by comparing the predicted natural frequencies $\hat{f}$ and mode shapes $\hat{\boldsymbol{\phi}}$ against their corresponding reference values $\overline{f}$ and $\overline{\boldsymbol{\phi}}$, which were evaluated via the FEM and are contained within $\mathbf{Y}_s^v$. Specifically, all the predicted frequencies $\hat{f}$ exhibit a coefficient of determination $R^2 > 0.99$ with respect to the corresponding FEM-based frequencies $\overline{f}$ for all considered modes. Moreover, the MAC values computed between each pair of predicted mode shapes $\hat{\boldsymbol{\phi}}$ and reference FEM mode shapes $\overline{\boldsymbol{\phi}}$ are consistently greater than 0.99. These results indicate an excellent training performance and confirm that the surrogate model $\hat{M}_s$ for the source structure accurately replicates its modal response with high fidelity.

\begin{figure}[t]
\centering
   \includegraphics[width = 1\textwidth]{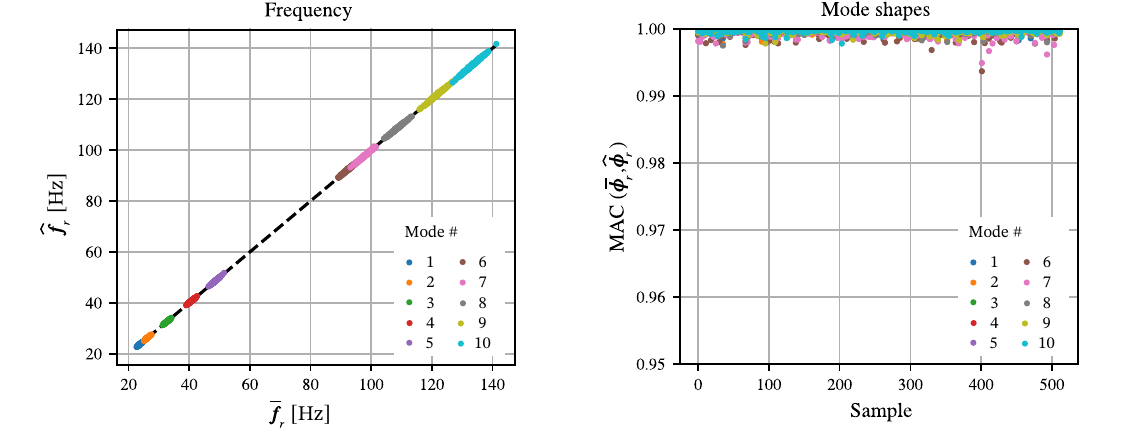}
\caption{Validation results of the surrogate model $\hat{M}_t$ obtained via transfer learning on the target beam model. Comparison between the predicted frequencies $\hat{f}$ and the FEM reference values $\overline{f}$, along with the MAC values computed between the predicted mode shapes $\hat{\phi}$ and the FEM mode shapes $\overline{\phi}$ for each sample of the validation set.}
    \label{fig:transfer_beams}
\end{figure}

The knowledge transfer from the source structure to the target structure is then performed, i.e., $\hat{M}_s \Rightarrow \hat{M}_t$. By freezing the general layers of the pre-trained FNN and keeping only the specialized layers trainable, the target training set $\left\{ \mathbf{X}_t, \mathbf{Y}_t \right\}$ is used to perform TL via fine-tuning over 1000 epochs, with the aim of constructing the SM for the target structure $\hat{M}_t$. The same custom loss function defined in Eq.~(\ref{eq.loss}) was employed, maintaining the weights $c_r, d_r$ for $r = 1, \dots, n$ equal to one, and the scaling factor $\beta$ equal to 100 - same parameters used in the training. Finally, a new validation set for the target structure, $\left\{ \mathbf{X}_t^v, \mathbf{Y}_t^v \right\}$, composed of $q_t^v = 512$ samples, was generated. The validation results of the target surrogate model $\hat{M}_t$ are reported in Figure~\ref{fig:transfer_beams}, showing that the MAC between predicted and FEM mode shapes, $\text{MAC}(\hat{\boldsymbol{\phi}}, \overline{\boldsymbol{\phi}})$, is consistently greater than 0.99, and the coefficient of determination $R^2$ for the natural frequencies exceeds 0.99. Therefore, the SM created via TL is deemed optimal for this theoretical case study.

% -------------------------------------
\subsection{Real case study: the Volumni Bridge and the M\'endez-N\'u\~nez Bridge}\label{section42}

\subsubsection{The Source Domain: The Volumni Bridge}\label{section421}

The Volumni Bridge is a five-spans prestressed concrete bridge located in Perugia, in the Italian region of Umbria. Designed by renowned Italian engineers Sergio and Vittorio Scalesse, the bridge was constructed in 1973 by ANAS S.p.A. . It serves as a critical link along the corridor between Rome and Florence, passing through the peri-urban area of Perugia.
The total length of the bridge is 340 meters, and the deck is composed of a continuous post-tensioned prestressed concrete multi-cell box girder. Span lengths are not uniform: the first and last spans measure 42.5 meters, while the three central spans extend to 85 meters each. The bridge is supported by four rectangular reinforced concrete piers. Pier heights vary according to the natural slope of the terrain, with the third pier being the tallest, reaching up to 12.8 meters. The adjacent piers are approximately 10 meters tall, whereas the first pier is significantly shorter. The piers are founded on deep pile foundations consisting of reinforced concrete piles.

% Model Calibration of the Volumni Bridge

\begin{figure}[H]
\centering
   \includegraphics[width=1\textwidth]{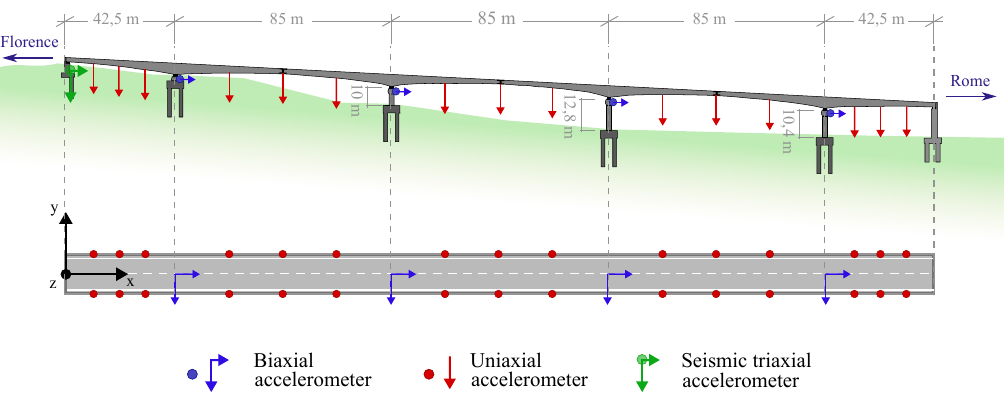}
   \caption{The Volumni Bridge. Sketch of the structure and sensor deployement.}    
    \label{fig:Volumni_bridge}
\end{figure}
\begin{figure}[t]
\centering
   \includegraphics[width=1\textwidth]{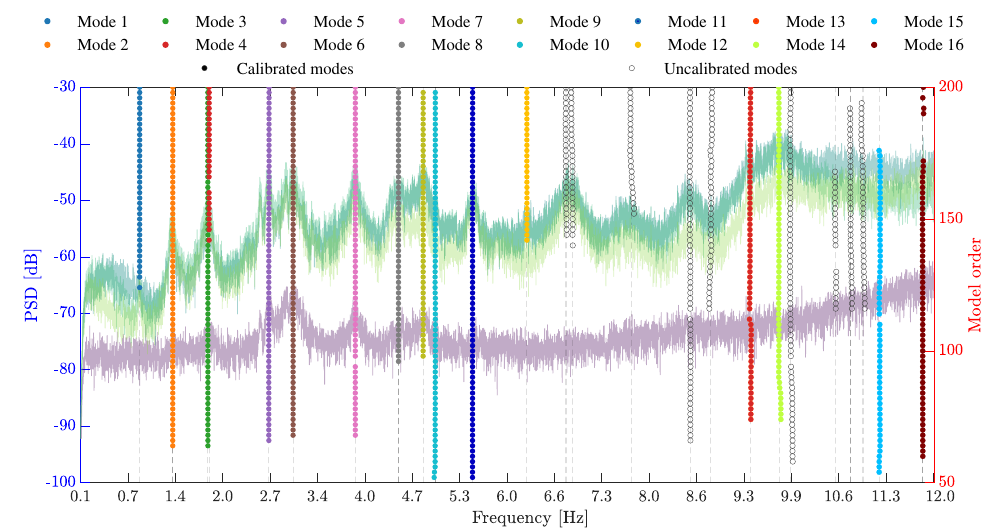}
   \caption{Stabilization diagram of the Volumni Bridge (24$^{\textrm{th}}$ April, 2024, 11:00 a.m.)}    
    \label{fig:Stab_Diag_Volumni}
\end{figure}
Within the \textit{SHM program} promoted by ANAS S.p.A.~\cite{ProgrammaSHM}, the Volumni Bridge was instrumented in 2023 with a comprehensive sensor network aimed at long-term structural monitoring, as depicted in Fig. \ref{fig:Volumni_bridge}. The dynamic monitoring system consists of one triaxial accelerometer, four biaxial accelerometers, and thirty uniaxial MEMS accelerometers, distributed strategically across the structure. Specifically, six uniaxial accelerometers were installed per span on the underside of the deck, while one biaxial accelerometer was positioned at the top of each pier and the triaxial accelerometer was placed at the base of the left abutment. The uniaxial and biaxial accelerometers are characterized by a measurement range of $\pm2$\,g, a sampling frequency of 4\,kHz, a dynamic range of 96\,dB, and a noise density of $25\,\mu\text{g}/\sqrt{\text{Hz}}$. The high-range triaxial seismic accelerometer was installed at the base of the left abutment, serving as a trigger in the event of seismic excitation. In total, the monitoring system comprises 41 accelerometric channels, providing a dense and reliable data stream for vibration-based analyses and long-term structural performance evaluation.

The FEM of the Volumni Bridge was developed based on original design drawings. The bridge deck was modeled as a continuous beam with variable cross-sections, incorporating 22 distinct sections across the three central spans and 11 sections across each of the lateral spans. To enable the extraction of modal displacements at DOFs corresponding to the sensor locations, rigid links and a flexible shells were introduced at the deck level. Due to the limitations of SAP2000 in directly accounting for torsional masses, these were included manually and calibrated individually for each variable cross-section to reflect the correct dynamic behavior. The bridge piers were modeled as prismatic beam elements with fixed supports at their bases, while the pier caps were similarly represented. Rigid links were used to connect the centroidal axis of the pier caps to their upper surfaces, where spring elements (with six DOFs each) were applied to simulate the sliding supports between the piers and the deck. The same modeling strategy was adopted for the sliding supports located at the abutments. This modeling approach ensures accurate representation of both the global stiffness distribution and the local dynamic response of the bridge, providing a reliable foundation for subsequent modal calibration and updating.

\begin{table}[H]
\setlength{\tabcolsep}{3pt} %% default is 6pt					
\newcommand\Tstrut{\rule{0pt}{0.3cm}}         % = `top' strut
\newcommand\Bstrut{\rule[-0.15cm]{0pt}{0pt}}   % = `bottom' strut	
 \footnotesize		
 \caption{Calibration report of the FEM of the Volumni Bridge.}
 \vspace{0.1cm}
   \centering
   \begin{tabular}{>{\centering\arraybackslash}m{1.5cm} 
    >{\centering\arraybackslash}m{1.5cm} 
    >{\centering\arraybackslash}m{2cm} 
    >{\centering\arraybackslash}m{2cm} 
    >{\centering\arraybackslash}m{0.1cm} 
    >{\centering\arraybackslash}m{1.5cm} 
    >{\centering\arraybackslash}m{1.5cm} 
}
   \toprule
   \multicolumn{7}{c}{\textbf{Calibration of the FEM of the Volumni Bridge}} \Tstrut\\
\midrule
\multirow{3}{*}{Mode} & \multicolumn{2}{c}{Experimental modes} & FEM & & \multirow{3}{*}{$\frac{\Delta f}{f_{exp}}$  [\%] }& \multirow{3}{*}{MAC  [-]} \Tstrut\\
\cmidrule(r){2-3} \cmidrule(r){4-5}
& {$f_{exp}$} [Hz] & $\xi_{exp}$ [\%] & $f_{FEM}$ [Hz] &  & & \Tstrut\\
\midrule
1 & 0.88 & 3.64 & 0.89 & & -1.40 & 0.95 \Tstrut\\
2 & 1.34 & 2.34 & 1.29 & & -4.62 & 0.90 \Tstrut\\
3 & 1.83 & 2.26 & 1.74 & & -4.73 & 0.88 \Tstrut\\
4 & 1.85 & 1.80 & 1.92 & & 2.55 & 0.84 \Tstrut\\
5 & 2.68 & 3.10 & 2.69 & & 2.39 & 0.88 \Tstrut\\
6 & 3.02 & 2.65 & 3.16 & & 4.87 & 0.86 \Tstrut\\
7 & 3.88 & 1.94 & 3.79 & & -3.81 & 0.90 \Tstrut\\
8 & 4.48 & 2.20 & 4.40 & & -2.03 & 0.85 \Tstrut\\
9 & 4.82 & 1.97 & 4.73 & & -0.16 & 0.80 \Tstrut\\
10 & 4.99 & 3.40 & 5.15 & & 3.92 & 0.83 \Tstrut\\
11 & 5.51 & 1.30 & 5.26 & & -4.70 & 0.92 \Tstrut\\
12 & 6.27 & 1.55 & 6.48 & & 3.48 & 0.77 \Tstrut\\
13 & 9.37 & 2.02 & 9.37 & & -3.67 & 0.83 \Tstrut\\
14 & 9.77 & 3.16 & 9.75 & & -3.33 & 0.93 \Tstrut\\
15 & 11.17 & 1.99 & 11.32 &  & 2.66 & 0.72 \Tstrut\\
16 & 11.77 & 1.54 & 11.63 &  &-0.48 & 0.80 \Tstrut\\
       \bottomrule
   \end{tabular}
   \label{tab:Volumni_Frequencies}
\end{table}

The FEM of the Volumni Bridge was calibrated using reference dynamic properties identified via the CoV-SSI method, as implemented in the MOVA/MOSS software~\cite{garcia2020mova}. Sixteen mode shapes were extracted from one hour of ambient vibration data recorded on April 24\textsuperscript{th}, 2024, at 11:00 a.m., with a sampling frequency of 100 Hz, and subsequently used as calibration targets. The identification procedure involved the computation of covariance matrices with a time lag of 2 seconds and the evaluation of model orders ranging from 50 to 200. Stable poles were selected based on tolerance thresholds of 1\% for natural frequencies and 2\% for MAC (Modal Assurance Criterion) values. Modes exhibiting damping ratios greater than 10\% were excluded, as were those with a Modal Phase Collinearity (MPC)~\cite{pappa1993consistent} index lower than 60\% or a Modal Phase Deviation (MPD)~\cite{dederichs2023experimental} exceeding 50\%. To isolate physical modes, a hierarchical clustering algorithm was applied, using a cut-off distance of 0.03 (defined as the sum of the relative frequency variation and 1–MAC) and a minimum cluster size of 20 poles. The resulting stabilization diagram, which highlights the consistency and reliability of the identified modal parameters, is shown in Figure~\ref{fig:Stab_Diag_Volumni}. Note that more than 16 modes were identified in the range between 0 and 12 Hz, but only 16 of them were fitted in the calibration process. The identified modes encompass both bending and torsional responses up to the third order, as well as a lateral bending mode (Mode 4). Notably, the lateral bending mode and the first-order torsional modes (Modes 10 and 11) are not global but instead appear to be localized primarily between spans 3 and 4. This spatial fragmentation of the modal response, particularly the absence of global torsional modes, is likely attributed to the differential height of the piers. In particular, the third and fourth spans are supported by the tallest piers, which result in relatively less rigid boundary conditions and high modal coupling between the piers and the adjacent piers and the supported spans.

\begin{figure}[H]
\centering
   \includegraphics[width=1\textwidth]{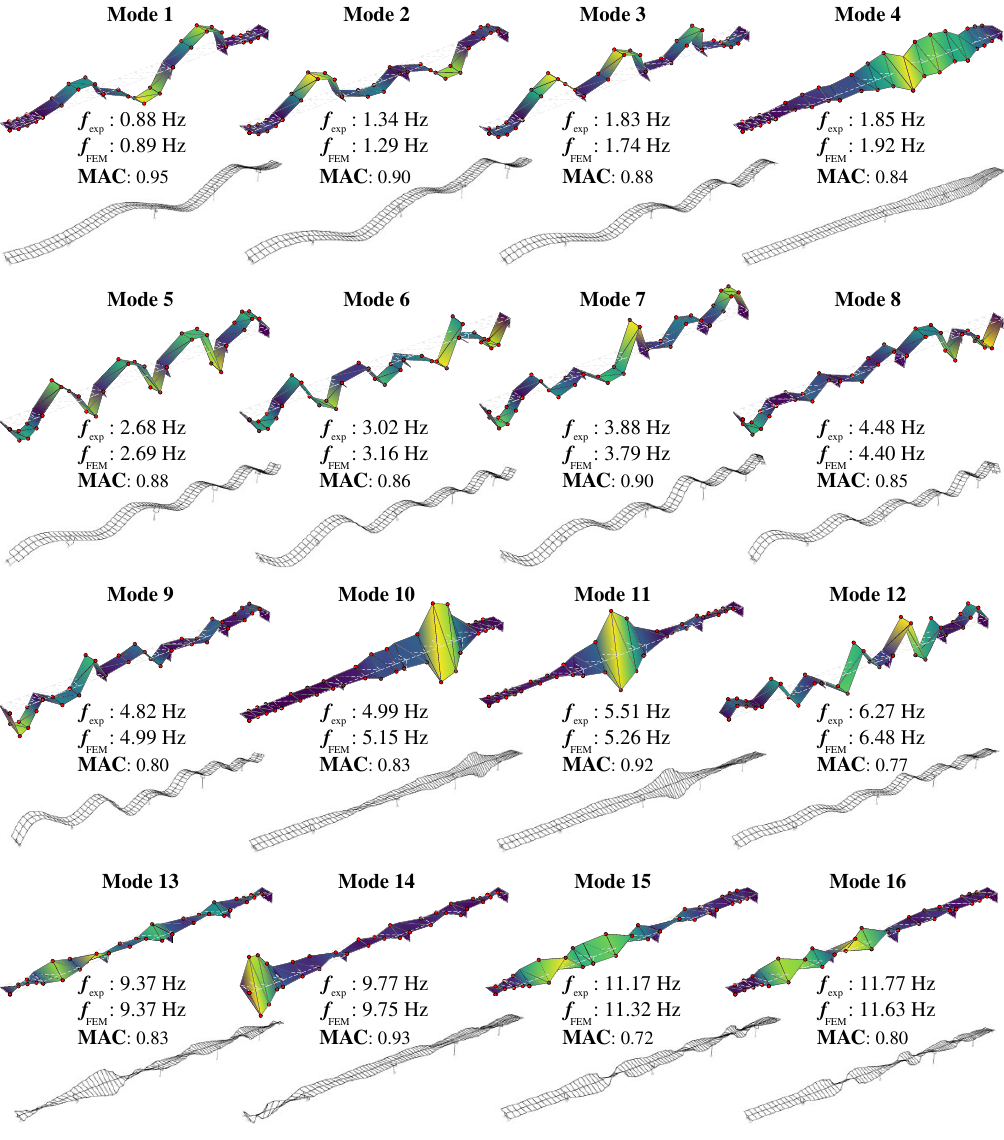}
   \caption{Calibrated modes of the Volumni Bridge.}    
    \label{fig:Calibrated_Volumni}
\end{figure}

The FEM calibration of the Volumni Bridge involved the optimization of 52 model parameters. These included: the elastic moduli of the deck and the piers (2 parameters), the mass densities of the deck and piers piers, and the mass density used to compute torsional masses of the beam (3 parameters), the stiffness components of the 6-degree-of-freedom springs modeling the supports at piers and abutments (6×6 = 36 parameters), and  scaling factors for the torsional inertia of the 11 distinct cross-sectional types along the deck (11 parameters). Following a sensitivity analysis, the support properties at Piers 2 and 4 were assumed to be equal, reducing the number of independent parameters to 46. Given the high dimensionality of the problem, a three-step calibration strategy was adopted to enhance convergence and robustness. This approach was necessary due to the large number of calibration parameters, which would otherwise lead to significant indeterminacy in the model updating process. In the first step, only flexural modes were considered by excluding the 12 parameters associated with torsional masses and the vertical and out-of-plane rotational stiffnesses of the supports, resulting in a reduced parameter set of 16. These parameters were calibrated independently. In the second step, the previously calibrated parameters were fixed, and the remaining variables—primarily associated with torsional behavior—were optimized to match the identified torsional modes. Finally, a joint optimization of all 46 parameters was performed to refine the entire model and improve overall agreement. All optimization stages were carried out using a Particle Swarm Optimization (PSO) \cite{EREIZ2022684} algorithm with a swarm size of 50, 50 and 100 particles and 80, 80, and 150 generations in the three stages, respectively. The objective function to be minimized was a weighted sum of the mean squared relative errors in natural frequencies and the dissimilarity in mode shapes, expressed as $1 - \text{MAC}$. A weighting factor of 2 was assigned to the frequency term and 1 to the mode shape term. To ensure robust comparison at each optimization step, an automatic mode-matching algorithm was adopted as implemented in the MOVA/MOSS framework \cite{garcia2020mova}. This function pairs numerical and experimental modes by minimizing a similarity cost function defined as a weighted sum of frequency difference and $1 - \text{MAC}$, using weights of 0.85 and 1, respectively. The final calibration results, including the relative frequency errors and MAC values, are summarized in Table~\ref{tab:Volumni_Frequencies}, while the corresponding mode shape comparisons are illustrated in Figure~\ref{fig:Calibrated_Volumni}.

The resulting calibrated model exhibited a strong agreement with the experimentally identified modal properties. Relative frequency errors for all 16 calibrated modes remained below 5\%, while MAC values exceeded 0.80 for the majority of modes. The only exceptions were Modes 12 and 15, which yielded MAC values of 0.77 and 0.72, respectively. The number of calibrated modes is relatively high, and the use of over 40 accelerometric channels significantly increases the difficulty of achieving very high MAC values since the numerical mode shapes must match a dense and spatially distributed set of experimental data points. Nevertheless, considering both the high number of modes and the overall accuracy of the frequency and mode shape correlations, the calibration can be deemed highly satisfactory. 
Overall, the results indicate a high degree of fidelity in reproducing the dynamic behavior of the Volumni Bridge. The calibrated FEM can thus be considered as a reliable digital twin of the structure, suitable for subsequent structural health assessment and the development of AI-based surrogate modeling approaches.

\subsubsection{The Target Domain: The M\'endez-N\'u\~nez Bridge}\label{section422}

\begin{figure}[H]
\centering
   \includegraphics[width=1\textwidth]{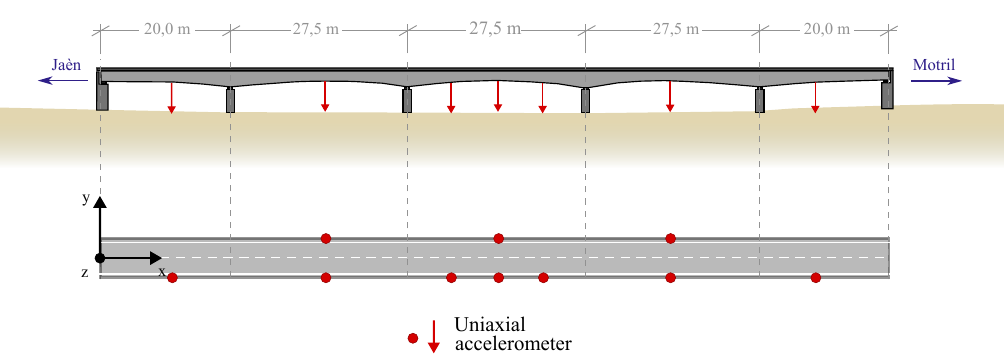}
   \caption{The  M\'endez-N\'u\~nez Bridge. Sketch of the bridge and sensor deployment.}    
    \label{fig:MN_bridge}
\end{figure}

The Méndez-Núñez Bridge is a continuous, five-span post-tensioned concrete bridge with a total length of 122.5 meters, located in the city of Granada, in the Autonomous Community of Andalucía, Spain (Fig.~\ref{fig:MN_bridge}). Constructed in March 1989 by the \textit{Dirección General de Carreteras} of the Province of Granada, the bridge facilitates vehicular traffic over Avenida de Andalucía, connecting the municipalities of Jaén and Motril. The deck consists of a variable-depth post-tensioned concrete box girder supported by four central piers made of reinforced concrete, with rectangular hollow cross-sections measuring 3.40 m $\times$ 1.50 m. These piers are founded on piles with pile caps. The structure includes two reinforced concrete abutments acting as retaining walls. All supports are equipped with elastomeric neoprene bearings of dimensions 0.90 m $\times$ 0.80 m $\times$ 0.15 m.

\begin{figure}[t]
\centering
   \includegraphics[width=1\textwidth]{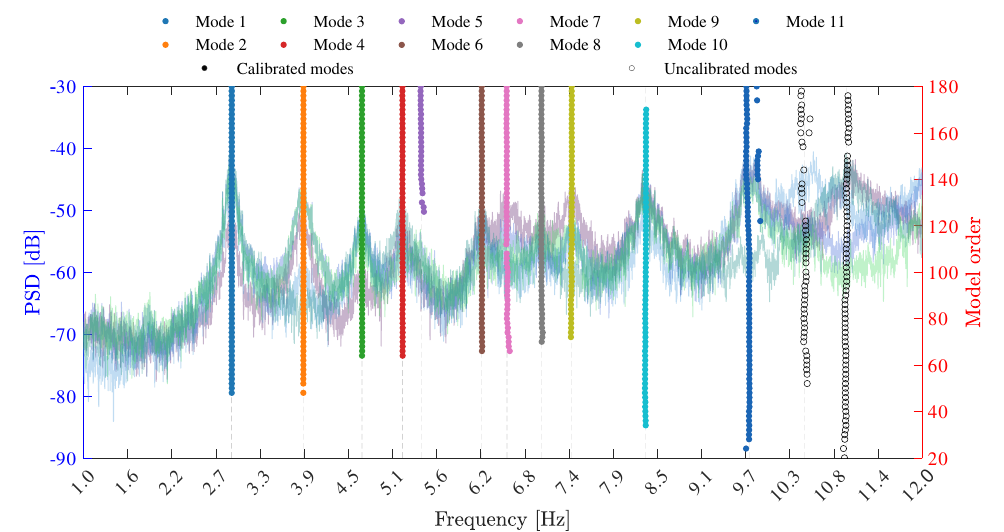}
   \caption{Stabilization diagram of the M\'endez-N\'u\~nez Bridge (30$^{th}$ November, 2023, 10:00 a.m.)}    
    \label{fig:Stab_Diag_MN}
\end{figure}

\begin{table}[H]
\setlength{\tabcolsep}{3pt} %% default is 6pt					
\newcommand\Tstrut{\rule{0pt}{0.3cm}}         % = `top' strut
\newcommand\Bstrut{\rule[-0.15cm]{0pt}{0pt}}   % = `bottom' strut	
 \footnotesize		
 \caption{Calibration report of the M\'endez-N\'u\~nez Bridge.}
 \vspace{0.1cm}
   \centering
   \begin{tabular}{>{\centering\arraybackslash}m{1.5cm} 
    >{\centering\arraybackslash}m{1.5cm} 
    >{\centering\arraybackslash}m{2cm} 
    >{\centering\arraybackslash}m{2cm} 
    >{\centering\arraybackslash}m{0.1cm} 
    >{\centering\arraybackslash}m{1.5cm} 
    >{\centering\arraybackslash}m{1.5cm} 
}
   \toprule
\multicolumn{7}{c}{\textbf{Calibration of the FEM of the M\'endez-N\'u\~nez Bridge}}  \Tstrut\\
\midrule
\multirow{3}{*}{Mode} & \multicolumn{2}{c}{Experimental modes} & FEM & & \multirow{3}{*}{$\frac{|\Delta f|}{f_{exp}}$  [\%] }& \multirow{3}{*}{MAC  [-]} \Tstrut\\
\cmidrule(r){2-3} \cmidrule(r){4-5}
& {$f_{exp}$} [Hz] & $\xi_{exp}$ [\%] & $f_{FEM}$ [Hz] &  & & \Tstrut\\
\midrule
1 & 2.87 & 1.19 & 2.87 & & 0.07 & 1.00 \Tstrut\\
2 & 3.76 & 1.19 & 3.79 & & 0.87 & 1.00 \Tstrut\\
3 & 4.53 & 1.86 & 4.61 & & 1.75 & 0.95 \Tstrut\\
4 & 5.05 & 1.29 & 5.07 & & 0.50 & 0.99 \Tstrut\\
5 & 5.31 & 2.50 & 5.22 & & 1.75 & 0.96 \Tstrut\\
6 & 6.14 & 1.28 & 6.13 & & 0.05 & 0.98 \Tstrut\\
7 & 6.42 & 2.12 & 6.42 & & 0.07 & 0.95 \Tstrut\\
8 & 6.81 & 2.50 & 6.93 & & 1.74 & 0.98 \Tstrut\\
9 & 7.24 & 1.43 & 7.23 & & 0.25 & 0.99 \Tstrut\\
10 & 8.31 & 1.39 & 8.33 & & 0.17 & 0.99 \Tstrut\\
11 & 9.51 & 1.57 & 9.67 & & 1.74 &  1.00 \Tstrut\\
       \bottomrule
   \end{tabular}
   \label{tab:MN_Frequencies}
\end{table}
As part of a national R\&D project, a permanent vibration-based SHM system was installed on September 27, 2023. The system comprises ten uniaxial piezoelectric accelerometers (model KB12VD; $\pm10\%$, 10.0 V/g, broadband resolution 1 $\mu$g RMS, $\pm$0.5 g pk), as shown in Fig.~\ref{fig:MN_bridge}. Ambient vibration data are collected in 30-minute intervals at a sampling frequency of 100 Hz, using a cDAQ-9184 acquisition system located on one of the piers. Additional environmental monitoring includes temperature data from four Pt1000/3850 sensors and humidity data via an AM2315 hygrometer. These sensors are managed by an Arduino Uno microcontroller and sampled every 5 minutes. OMA is performed using MOVA/MOSS software installed on a local edge device, and identified modal properties are automatically transmitted online for remote visualization. Raw time series are stored locally and periodically collected.

%Finite Element Modeling

A three-dimensional FEM of the bridge was developed in SAP2000 based on structural drawings and in-situ inspections. The bridge deck was modeled using frame elements with variable cross-sections and lumped torsional masses. The piers were represented as frame elements with tapered cross-sections and were connected to the deck via longitudinal and vertical springs whose stiffness values were derived from the bearing pad properties and Spanish standards. Additional massless, rigid elements were included to model sensor locations and eccentricities between the deck and substructure.

% Modal Identification and Model Calibration
The dynamic properties of the structure were identified using the  CoV-SSI method. Covariance matrices were generated using a time lag of 3.2 seconds, and a stabilization diagram was constructed by extracting poles over model orders ranging from 2 to 180. To ensure the selection of reliable and consistent modal parameters, strict selection tolerances were applied: 0.1\% for natural frequencies, 0.5\% for damping ratios, and 0.005 for MAC values. 

\begin{figure}[H]
\centering
   \includegraphics[width=1\textwidth]{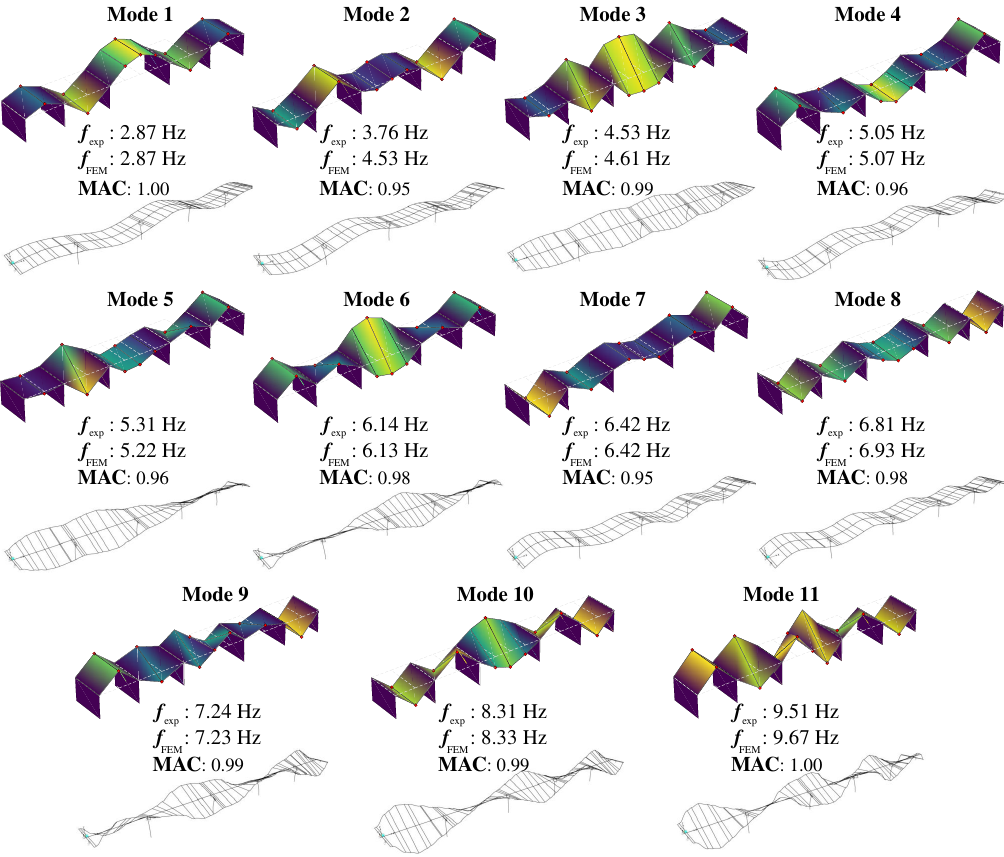}
   \caption{Calibrated mode shapes of the M\'endez-N\'u\~nez Bridge.}    
    \label{fig:Calibrated_MN}
\end{figure}

Physical modes were subsequently extracted through a hierarchical clustering algorithm, adopting a cut-off distance of 0.01 and a minimum cluster size of 20 poles. This process led to the robust identification of eleven global vibration modes with frequencies below 12 Hz. The identified modes comprise five flexural modes and six torsional modes, providing a rich dataset for model calibration. These modes are illustrated in Figure~\ref{fig:Calibrated_MN}.

For the purpose of model updating, the bridge deck was discretized into eleven segments, each characterized by an independent elastic modulus to account for potential spatial variability in stiffness. A parameter optimization procedure was carried out using a  PSO algorithm, involving 60 particles over 65 generations. The objective function was defined as a weighted sum of the maximum relative frequency error and the mean of $1 - \text{MAC}$ across all identified modes. A weighting factor of 2 was assigned to the frequency discrepancy term, while the mode shape term was assigned a factor of 1, emphasizing the prioritization of frequency accuracy in the calibration process.

As a result of this calibration process, a total of eleven global modes —both flexural and torsional—were successfully matched and aligned between the experimental data and the numerical model. Table~\ref{tab:MN_Frequencies} summarizes the comparison between the experimental and simulated modal parameters, including relative frequency errors and MAC values, while Figure~\ref{fig:Calibrated_MN} presents a visual comparison of the identified and calibrated mode shapes. The achieved accuracy in both frequency and mode shape correlation demonstrates the effectiveness of the adopted calibration methodology and confirms the reliability of the resulting numerical model for subsequent analyses.

%%%%%%%%%%%%%%%%%%%%%%%%%%%%%%%%%%%%%%%%%%%%%%%%%%%%%%%%%
\subsubsection{Training of the Neural Network on the source bridge}\label{section423}

\begin{figure}[H]
\centering
   \includegraphics[width = 1\textwidth]{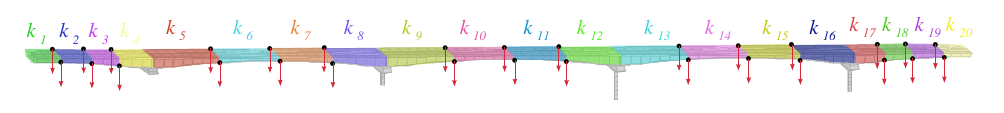}
   \caption{Control parameters defined on the Volumni Bridge and selected FEM and sensor nodes. Black circles denote DoFs selected for mode shape extraction in the FEM. Red arrows indicate the locations of the accelerometric channels installed as part of the SHM system.}  
    \label{fig:Volumni_parameters}
\end{figure}

To develop the model-driven SM capable of predicting the dynamic modal properties of the Volumni Bridge, a synthetic training dataset was generated using the calibrated high-fidelity FEM described in Section~\ref{section421}. The FEM was parametrized by discretizing the deck into $N = 20$ control regions, uniformly distributed along the five spans of the bridge, with each span divided into four segments of length equal to a quarter of the span length as depicted in Fig. \ref{fig:Volumni_parameters}. This resulted in 20 stiffness multipliers $k_i$, with $i = 1, \dots, N$, each representing a scaling factor of the nominal Young’s modulus for a specific region of the deck. These stiffness multipliers  $k_i$ were assumed to vary independently within the bounded interval $[a_i, b_i] = [0.80,\,1.05]$ for all $i=1, \dots , N$, accounting for plausible spatial fluctuations in material properties. Note that the limit cases of $k_i=0.80 \, \forall i$, $k_i=1.00 \, \forall i$, and $k_i=1.05 \, \forall i$ represent, respectively, a global reduction of 20\% in the elastic moduli, the undamaged (healthy) condition, and a global increase of 5\%. The upper bound is set above 1.00 to prevent solutions in the subsequent damage identification from concentrating at the boundary, which could lead to convergence difficulties. It also serves to accommodate benign fluctuations induced by environmental conditions.

\begin{figure}[H]
\centering
   \includegraphics[width = 1\textwidth]{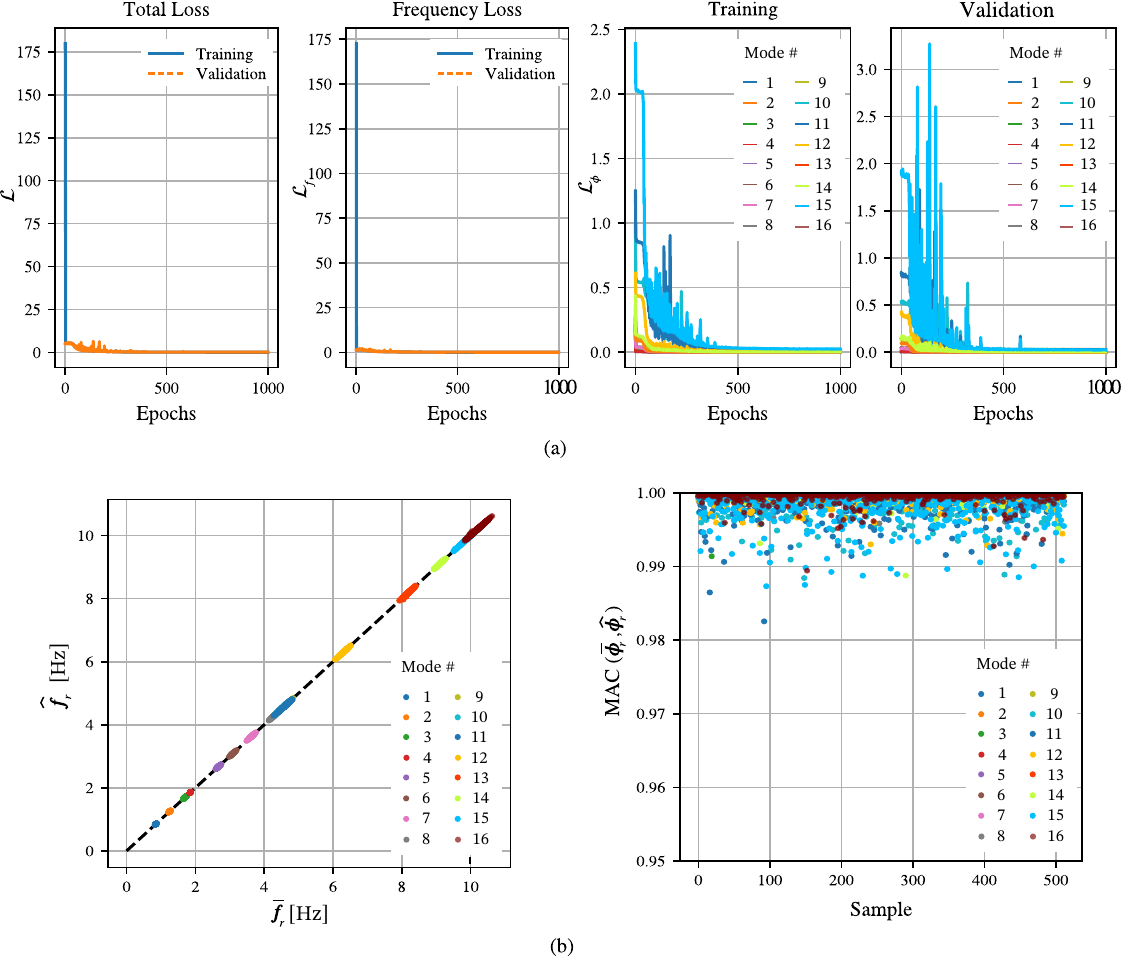}
   \caption{Training of the surrogate model $\hat{M}$ on the Volumni Bridge. (a) Total loss and individual losses associated with frequency and mode shape predictions during the training phase. (b) Comparison between the predicted frequencies $\hat{f}$ and the FEM reference values $\overline{f}$, along with the MAC values computed between the predicted mode shapes $\hat{\phi}$ and the FEM mode shapes $\overline{\phi}$ for each sample of test set.}    
    \label{fig:training_Volumni}
\end{figure}

To sample the parameter space efficiently and ensure good coverage, an LHS strategy was adopted. A training set of $q_s = 1024$ samples was generated, with each sample consisting of a unique combination of the $k_i$ multipliers. For each sampled configuration, a corresponding modal analysis was performed using the calibrated FEM. This yielded a set of numerical outputs $\mathbf{Y}_s$, consisting of resonant frequencies and mode shapes associated with each configuration. 
Given the objective of developing a SM specifically sensitive to damage affecting the bridge deck, only the DOFs of the FEM corresponding to the locations of the deck-mounted sensors were considered. Consequently, the mode shapes included in the output dataset $\mathbf{Y}_s$ were projected onto this reduced set of DOFs, resulting in $m = 36$ modal displacements corresponding to the locations of the uniaxial accelerometers installed beneath the girder along the five spans of the bridge, as illustrated in Figure~\ref{fig:Volumni_parameters}. This selection ensures full coherence between the numerical model and the spatial resolution of the SHM system, improving the physical relevance of the modal features and enhancing the sensitivity of the SM to local variations in deck stiffness.
The full training dataset $\{ \mathbf{X}_s, \mathbf{Y}_s \}$ thus captures the functional dependency between the non-dimensional damage-sensitive parameters (i.e., $\mathbf{X}_s$, derived from the stiffness multipliers) and the modal properties of the bridge.

%Neural Network Architecture and Training

The network architecture depicted in Section \ref{section31} is structured as follows for the specific case:

\begin{itemize}
    \item \textbf{Input layer:} A fully connected layer of dimension $N=20$, corresponding to the number of structural parameters $\boldsymbol{\pi}$ inside the design space $\mathbf{X}_s$.

    \item \textbf{General layers:} It is structured as a the shared feature extractor. The input is processed through a sequence of 5 fully connected layers with hyperbolic tangent activation functions. The dimensionality is progressively expanded from 64 to 512 neurons. This shared representation encodes the input parameters into a latent space suitable for both frequency and mode shape prediction.

    \item \textbf{Frequency prediction branch:} This branch estimates the natural frequencies of the structure:
    \begin{itemize}
        \item The shared latent vector is processed through two dense layers with 512 neurons each, followed by dropout layers with a dropout rate of 0.2.
        \item Subsequently, three dense layers with GELU activation function and 256 neurons are applied.
        \item The final output layer consists of $n=16$ neurons, one per predicted mode frequency
    \end{itemize}

    \item \textbf{Mode shape prediction branches:} Each mode shape is predicted by a dedicated branch:
    \begin{itemize}
        \item Each branch receives the output of the last layer of the general layers.
        \item The signal is passed through a stack of 2 to 6 fully connected layers, typically composed of 256 or 128 neurons.
        \item Some layers incorporate dropout regularization with a rate of 0.3 to assure a better generalization.
        \item The final layer contains $m = 41$ neurons, representing the spatial discretization of the mode shape at sensor locations. The activation function is the hyperbolic tangent.
    \end{itemize}
\end{itemize}

The FNN was trained over 1000 epochs using the training population $\{ \mathbf{X}_s, \mathbf{Y}_s \}$ in order to construct the surrogate model $\hat{M}_s$ with 15\% of the samples reserved for internal validation. The custom loss function, as defined in Eq.~(\ref{eq.loss}), was employed to jointly assess the accuracy of predicted modal frequencies and mode shapes. Specifically, the frequency weights $c_i$ were set to 1 for all $i \in [1, 20]$, while the mode shape weights $d_i$ were set to 1 for all modes except for mode 15 and 16, for which $d_{15} = d_{16} = 0.80$ were used to reduce the higher fluctuations that generate in the loss, that were found to reduce the capability of the FNN to generalize and converge. The scaling parameter $\beta$ was set to 1000 to balance the relative contributions of frequency and mode shape terms. The neural architecture's hyperparameters were tuned in order to balance the convergence behavior of the training and validation loss curves against the overall predictive performance on both frequencies and mode shapes. The network was optimized using the Adam optimizer, with an initial learning rate of $2 \cdot 10^{-3}$ and a momentum coefficient of 0.98. A super mini-batch size of 4 was used, and data shuffling was applied at each epoch to enhance convergence and generalization. The training history, illustrated in Figure~\ref{fig:training_Volumni}(a), shows the loss function evolution over epochs for both training and validation, with the individual contributions of frequency and mode shape errors clearly separated. Stabilization was generally achieved after approximately 500 epochs. 

To evaluate the generalization capabilities of the SM, an independent validation dataset $\{ \mathbf{X}_s^v, \mathbf{Y}_s^v \}$ of $q^v_s = 512$ samples was also generated following the same LHS-based approach.  Figure~\ref{fig:training_Volumni}(b) compares the predicted frequencies $\hat{f}$ with those from the FEM $\overline{f}$, and shows the MAC values between predicted mode shapes $\hat{\boldsymbol{\phi}}$ and FEM references $\overline{\boldsymbol{\phi}}$. For all 16 modes, the coefficient of determination $R^2$ exceeded 99\%, and MAC values remained consistently above 0.98, confirming the surrogate's high fidelity in reproducing the modal response.

\subsubsection{Transfer Learning}\label{section425}

TL was implemented by leveraging the trained FNN surrogate model $\hat{M}_s$ developed for the Volumni Bridge (source structure) to construct the surrogate model $\hat{M}_t$ for the M\'endez-N\'u\~nez Bridge (target structure). The goal was to transfer modal knowledge between the two structures to enable efficient learning on the target, despite a reduced availability of training data.

\begin{figure}[H]
\centering
   \includegraphics[width = 1\textwidth]{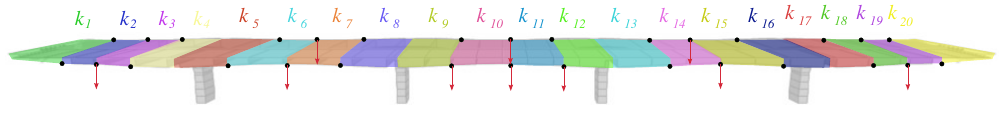}
   \caption{Control parameters defined on the M\'endez-N\'u\~nez Bridge and selected FEM and sensor nodes. Black circles denote DOFs selected for mode shape extraction in the FEM. Red arrows indicate the actual locations of the accelerometric channels installed as part of the SHM system.}  
    \label{fig:MN_parameters}
\end{figure}

\begin{table}[H]
\setlength{\tabcolsep}{3pt} %% default is 6pt					
\newcommand\Tstrut{\rule{0pt}{0.3cm}}         % = `top' strut
\newcommand\Bstrut{\rule[-0.15cm]{0pt}{0pt}}   % = `bottom' strut	
 \footnotesize		
 \caption{Transferred mode shapes of the Volumni Bridge and of the M\'endez-N\'u\~nez Bridge.}
 \vspace{0.1cm}
   \centering
   \begin{tabular}{>{\centering\arraybackslash}m{1.5cm}
    >{\centering\arraybackslash}m{0.1cm}
    >{\centering\arraybackslash}m{2cm} 
    >{\centering\arraybackslash}m{1.5cm} 
    >{\centering\arraybackslash}m{0.1cm}  
    >{\centering\arraybackslash}m{2cm} 
    >{\centering\arraybackslash}m{1.5cm}
    >{\centering\arraybackslash}m{0.1cm}
    >{\centering\arraybackslash}m{0.8cm}
    >{\centering\arraybackslash}m{0.8cm}
    >{\centering\arraybackslash}m{0.1cm}
    >{\centering\arraybackslash}m{3.4cm}
}
   \toprule
\multicolumn{12}{c}{\textbf{Transferred modes}}  \Tstrut\\
\midrule
\multirow{2}{*}{Transfer} & & \multicolumn{2}{c}{Volumni} & & \multicolumn{2}{c}{M\'endez-N\'u\~nez} & & \multirow{3}{*}{$\Delta f_k$ [Hz]} & \multirow{3}{*}{MAC$_k$ [-]} & & \multirow{3}{*}{Description} \Tstrut\\
\cmidrule(r){3-4} \cmidrule(l){6-7}
 mode $k$ & & Source Mode & {$f_{exp}$} [Hz] & & Target Mode & $f_{exp}$ [Hz] & & & & & \Tstrut\\
\midrule
1 & & 1 &  0.88 & & 1 &  2.87 & & -1.99 & 0.95 & & 1st order bending mode \Tstrut\\
2 & & 2 &  1.34 & & 2 &  3.76 & & -2.42 & 0.87 & & 1st order bending mode \Tstrut\\
3 & & 3 &  1.83 & & 4 &  5.05 & & -3.22 & 0.75 & & 1st order bending mode \Tstrut\\
4 & & 4 &  1.85 & & 3 &  4.53 & &-2.68  & 0.78 & & Lateral mode \Tstrut\\
5 & & 5 &  2.68 & & 7 &  6.42 & & -3.74& 0.57 & & 2nd order bending mode \Tstrut\\
6 & & 6 &  3.02 & & 8 &  6.81 & & -3.79& 0.85 &  &  2nd order bending mode \Tstrut\\
7 & & 10 &  4.99 & & 5 &  5.31 & & -0.32 & 0.17 &  & 1st order torsional mode \Tstrut\\
8 & & 11 & 5.51 &  & 6 &  6.14 & & -0.63 & 0.73 & & 1st order torsional mode \Tstrut\\
       \bottomrule
   \end{tabular}
   \label{tab:Transferred_Modes}
\end{table}

\begin{figure}[t]
\centering
   \includegraphics[width=1\textwidth]{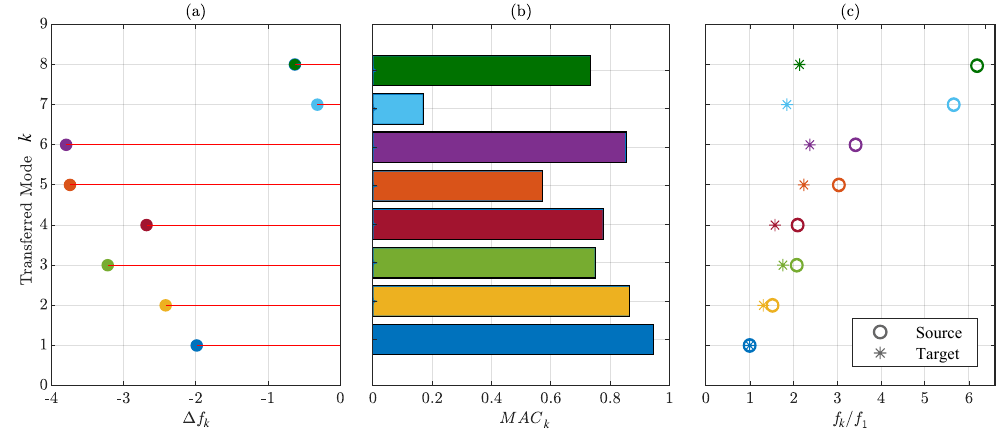}
   \caption{Key characteristics of the modes selected for transfer learning. (a) Relative frequency differences $\Delta f_k$
  between each pair of matched modes and (b) corresponding MAC$_k$ values. (c) Comparison between the distributions of modal frequencies, normalized by their respective fundamental frequencies.}    
    \label{fig:Transfered_modes_plots}
\end{figure}

The first step involved identifying a subset of modes to transfer. This selection was based on the similarity in terms of MAC and typological similarity between the real reference mode shapes that have been calibrated on the FEMs. As summarized in Table~\ref{tab:Transferred_Modes}, a total of 8 modes were selected, exhibiting high mutual similarity either in terms of MAC values or shared modal order and deformation type (e.g., first bending, first torsion, etc.). Figure~\ref{fig:Transfered_modes_plots} provides further insight into the modes selected to be transferred. Panel (a) reports the relative frequency differences $\Delta f_k, \; k=1, \dots, 8$ between the paired modes while panel (b) shows the MAC values for each matched pair. Finally, panel (c) compares the distribution of modal frequencies, normalized by the respective fundamental frequency, to illustrate their relative positions in the dynamic spectrum.
While most transferred modes exhibit satisfactory modal correlation, Mode 7 shows a notably lower MAC value (MAC$_7$ = 0.17). This discrepancy is attributed to structural differences: the torsional modes of the Volumni Bridge tend to be spatially localized due to the irregular pile height distribution, whereas the M\'endez-N\'u\~nez Bridge features uniform pier heights, promoting global torsional modes. Despite this, Transfer Mode 7 was retained in the transfer set to test the network’s ability to generalize beyond structural-specific boundary conditions and reflect a broader dynamic behavior. This mode selection process establishes the basis for effective knowledge transfer and cross-structure generalization of the SM, as discussed in Section~\ref{section32}.

The FEM of the M\'endez-N\'u\~nez Bridge was parameterized using the same discretization scheme adopted for the Volumni Bridge. Each $k_i$ was assumed to vary independently within the interval $[ a_i, b_i] =[0.80,1.05]$ and a LHS strategy was employed to sample the high-dimensional parameter space. This resulted in a design dataset ${ \mathbf{X}_t, \mathbf{Y}_t }$ of $q_t = 256$ samples. In order to ensure maximum informational consistency and compatibility with the source model, instead of limiting the selection to the sensor layout of the M\'endez-N\'u\~nez Bridge, all nodal positions corresponding to accelerometer locations in the Volumni Bridge were used. As a result, the mode shape vectors $\overline{\boldsymbol{\phi}}_{r,k}$ included in $\mathbf{Y}_t$ were composed of $m = 36$ components (instead of 10), thus increasing the information content in the output and consistently with the source domain representation (see Figure~\ref{fig:MN_parameters}). The resulting dataset captures the functional relationship between the non-dimensional damage-sensitive parameters $\mathbf{X}_t$ and the modal features $\mathbf{Y}_t$ for the target structure. To comprehensively assess the generalization ability of the transferred model, an independent and significantly larger validation dataset ${ \mathbf{X}_t^v, \mathbf{Y}_t^v }$ of size $q_t^v = 1024$ samples was generated using the same LHS-based methodology. The relatively small size of the training dataset was a deliberate choice aimed at avoiding overfitting and preserving the pretrained knowledge in the specialized layers of the source model. Conversely, the extended validation dataset allowed for a robust and extensive evaluation of the generalization performance of the transferred SM.

\begin{figure}[t]
\centering
   \includegraphics[width = 1\textwidth]{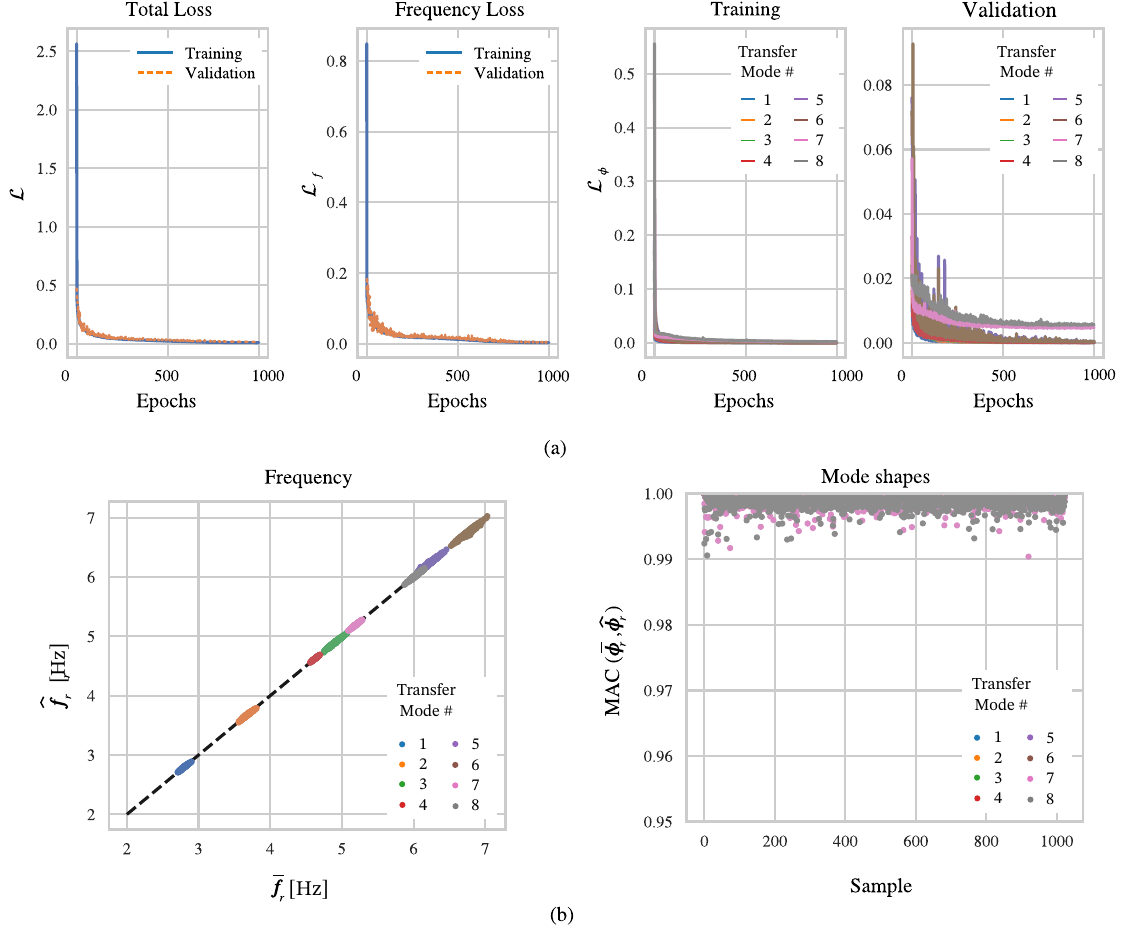}
   \caption{Transfer learning via fine tuning of the surrogate model $\hat{M}_s$ of the Volumni Bridge to build the $\hat{M}_t$  of the M\'endez-N\'u\~nez Bridge. (a) Total loss and individual losses associated with frequency and mode shape predictions during the fine tuning. (b) Comparison between the predicted frequencies $\hat{f}$ and the FEM reference values $\overline{f}$, along with the MAC values computed between the predicted mode shapes $\hat{\phi}$ and the FEM mode shapes $\overline{\phi}$ for each sample of the validation set.}    
    \label{fig:transfer_MN}
\end{figure}

The TL process was implemented through fine-tuning, starting from the source neural network surrogate model $\hat{M}_s$ trained on the Volumni Bridge, and adapting it to the target structure, the M\'endez-N\'u\~nez Bridge. The general feature extraction layers  were frozen, while only the specialized output branches—dedicated to modal frequencies and shapes—were retrained. The fine-tuning phase lasted for 1000 epochs using the target training dataset ${ \mathbf{X}_t, \mathbf{Y}_t }$, with 20\% of the samples reserved for internal validation. To enable the transfer of only the 8 selected modes, the output layer of the frequency branch was modified by reducing the number of neurons from 16 to 8. Simultaneously, the loss function defined in Eq.~(\ref{eq.loss}) considers only the modes that had been selected for transfer. Therefore, the custom loss function defined in Eq.~(\ref{eq.loss}) was retained, with frequency and mode shape weights $c_r$ and $d_r$ set to 1 for $r \in [1, 8] \cup [10, 11]$ and 0 for $r \in [7, 9]$ (accordingly to Table \ref{tab:Transferred_Modes}). To mitigate excessive fluctuations in the early stages of training, the scaling parameter $\beta$ was reduced to 2. The same mini-batch size of 4 was adopted, with data shuffling applied at every epoch to promote convergence and enhance generalization. The results in terms of training and validations are reported in Figs. \ref{fig:transfer_MN}(a) and \ref{fig:transfer_MN}(b), respectively. It can be noted that for all 8 transferred modes, the coefficient of determination $R^2$ exceeded 99\%, and MAC values remained consistently above 0.99, confirming the optimal TL results and assessing the effectiveness of the proposed methodology.

\subsubsection{Bayesian-based model updating of the target bridge}\label{section325}

\begin{figure}[H]
\centering
   \includegraphics[width=1\textwidth]{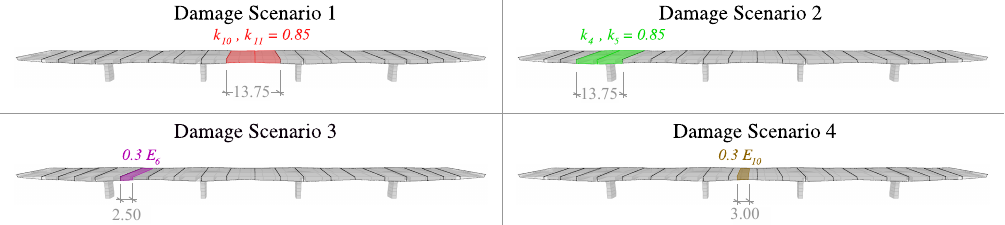}
   \caption{Synthetic damage scenarios defined using the FEM of the M\'endez-N\'u\~nez Bridge (dimensions in m).}    
    \label{fig:Damage_scenarios}
\end{figure}

To assess the effectiveness of the SM-based Bayesian damage assessment strategy proposed in Section \ref{section22}, four synthetic damage scenarios were defined on the FEM of the M\'endez-N\'u\~nez Bridge, designed to emulate realistic deterioration mechanisms such as localized cracking or distributed deck degradation. These scenarios, illustrated in Figure~\ref{fig:Damage_scenarios}, were selected to test the SM-based Bayesian inference framework's ability to detect and quantify structural damage under varying conditions of spatial symmetry, localization, and intensity. Specifically, the methodology is tested in the following contexts:
\begin{itemize}
    \item \textbf{Damage affecting the full extent of the control region:} In this case, a damage intensity which falls within the boundaries defined as $a_i \in [0.80,\,1.05], \; i = 1, \dots, N$ and $b_i \in [0.80,\,1.05], \; i = 1, \dots, N$.
    \begin{description}
        \item[\textit{Damage Scenario 1:}] This scenario aims to assess the model's ability to detect and quantify damage in a perfectly symmetric location. Given the global nature of the mode shapes and the absence of localized modes in this area, combined with the structural symmetry of the bridge, damage in this location might be more challenging to detect. The scenario involves a 15\% reduction in stiffness across the entire length of the control regions governed by $k_{10}$ and $k_{11}$, located in the middle of the central span, i.e., $k_{10} = k_{11} = 0.85$.
        \item[\textit{Damage Scenario 2:}] This scenario targets an asymmetric damage configuration to test whether the model can still accurately infer damage in a non-symmetric location. The damage consists of a 15\% stiffness reduction in regions governed by stiffness multipliers $k_{4}$ and $k_{5}$, positioned on the left and on the right of the first pier. Both parameters are set to $k_{4} = k_{5} = 0.85$.
    \end{description}
    \item \textbf{Damage affecting a partial extent of the control region:} In this case, the objective is to test the ability to detect more severe and localized damages outside the predefined boundaries, which could resemble phenomena such as flexural cracking (the elastic modulus directly influences the local bending flexural stiffness) or shear cracking (the elastic modulus affects the shear stiffness modulus $G$).
    \begin{description}
        \item[\textit{Damage Scenario 3:}] This scenario examines the model's performance in identifying partial damage located in a non-symmetric position, affecting less than half of the control region's length. The damage is applied within the region governed by $k_{6}$, situated in the second quarter of the second span. The total control region length is 6.875 m, while the damage is confined to a 2.5 m long segment on the left side. The local stiffness is reduced to 30\% of its nominal value, i.e., $0.3 \times E_6$.
        \item[\textit{Damage Scenario 4:}] This case investigates localized damage in a nearly symmetric location. The affected region is governed by $k_{10}$, located in the second quarter of the central span. Within a total length of 6.875 m, the damage is applied to a 3 m central portion. As in the previous case, the stiffness is reduced to 30\% of the nominal value, that is $0.3 \times E_{10}$.
    \end{description}
\end{itemize}

\begin{table}[H]
\setlength{\tabcolsep}{3pt} %% default is 6pt					
\newcommand\Tstrut{\rule{0pt}{0.3cm}}         % = `top' strut
\newcommand\Bstrut{\rule[-0.15cm]{0pt}{0pt}}   % = `bottom' strut	
 \footnotesize		
 \caption{Damage-induced decays in the resonant frequencies and MAC values (between undamanged and damaged mode shapes) of the M\'endez-N\'u\~nez Bridge under damage scenarios 1 to 4.}
 \vspace{0.1cm}
   \centering
   \begin{tabular}{c c c c c c c c c}
   \toprule
& \multicolumn{8}{c}{\textbf{Frequency decays} [\%]}\Tstrut\\
\cmidrule(r){2-9} 
Case scenario & Mode 1 & Mode 2 & Mode 3 & Mode 4 & Mode 5 & Mode 6 & Mode 7 & Mode 8 \Tstrut\\
\midrule
Damage Scenario 1 & -1.91 & -0.42 & -0.24 & -1.01 & -0.42 & -0.41 & -0.25 & -0.39 \Tstrut\\
Damage Scenario 2 & -0.26 & -0.18 & -0.34 & -0.38 & -0.42 & -0.25 & -1.61 & -1.01 \Tstrut\\
Damage Scenario 3 & -0.67 & -1.92 & -1.39 & -1.83 & -0.92 & -0.06 & -0.53 & -0.15 \Tstrut\\
Damage Scenario 4 & -4.75 & -0.78 & -0.63 & -2.32 & -1.08 & -1.04 & -0.50 & -0.80 \Tstrut\\
\midrule
& \multicolumn{8}{c}{\textbf{MACs} [\%]}\Tstrut\\
\cmidrule(r){2-9} 
Case scenario & Mode 1 & Mode 2 & Mode 3 & Mode 4 & Mode 5 & Mode 6 & Mode 7 & Mode 8 \Tstrut\\
\midrule
Damage Scenario 1 & 99.93 & 99.99 & 99.96 & 99.91 & 99.96 & 99.91 & 99.98 & 99.97 \Tstrut\\
Damage Scenario 2 & 99.99 & 99.99 & 99.91 & 99.90 & 99.83 & 99.93 & 96.63 & 98.49 \Tstrut\\
Damage Scenario 3 & 99.88 & 99.23 & 98.64 & 99.01 & 98.75 & 99.91 & 99.56 & 99.98 \Tstrut\\
Damage Scenario 4 & 99.39 & 99.43 & 99.41 & 99.15 & 98.50 & 98.85 & 99.32 & 99.30 \Tstrut\\
       \bottomrule
   \end{tabular}
   \label{tab:frequency_decays}
\end{table}

As previously described, each scenario was implemented by reducing the stiffness in selected segments of the FEM, thereby simulating damage-induced degradation of the deck’s structural capacity. Subsequently, linear modal analyses were carried out for each scenario to evaluate the resulting variations in both natural frequencies and mode shapes. The frequency variations, expressed as percentage relative decays, and the MAC values between the damaged and undamaged mode shapes are summarized in Table~\ref{tab:frequency_decays} for the modes on which the surrogate model $\hat{M}_t$ of the M\'endez-N\'u\~nez Bridge is defined.

\begin{figure}[H]
\centering
   \includegraphics[width=1\textwidth]{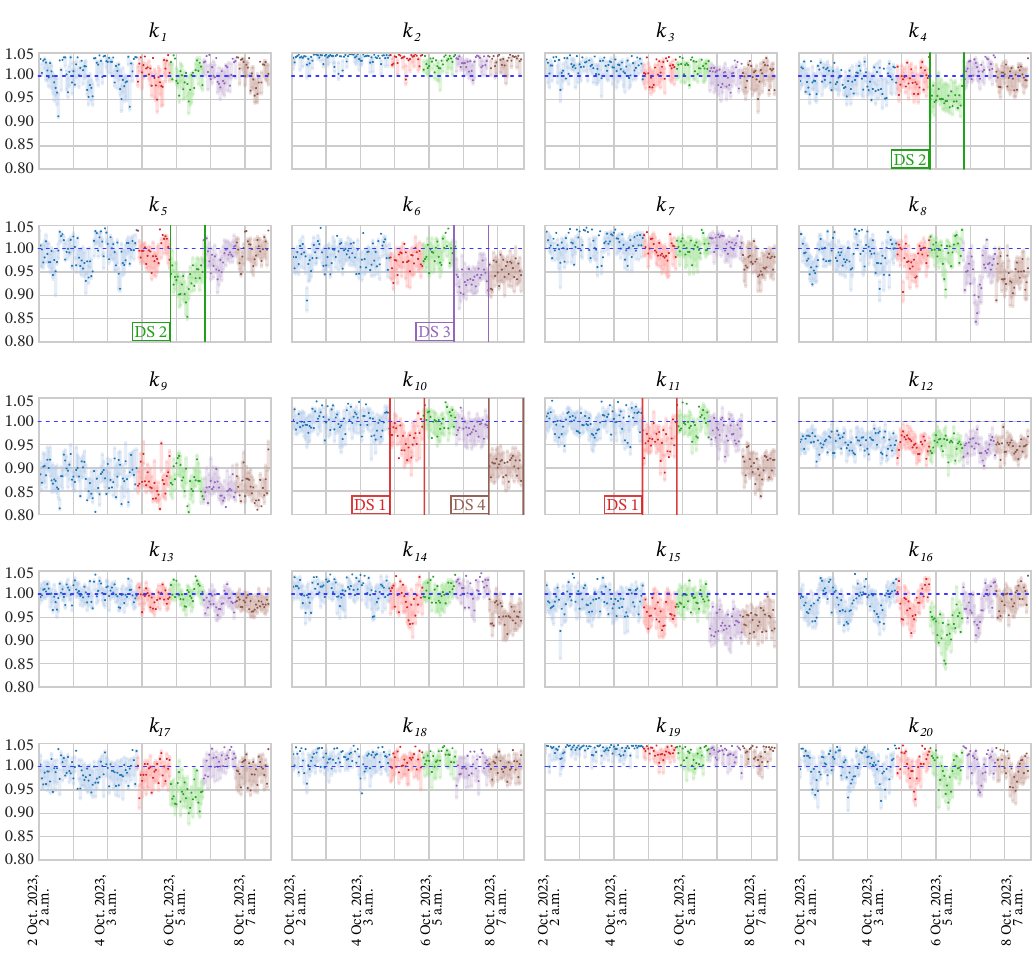}
   \caption{Median and interquartile range (25th–75th percentiles) of the posterior distributions during the period from October 2 to October 9, 2023, under the imposed damage scenarios accounting only for frequency predictions in the likelihood. Blue represent the undamaged configuration; red corresponds to Damage Scenario 1, green to Scenario 2, violet to Scenario 3, and brown to Scenario 4.}    
    \label{fig:Posterior_means_std}
\end{figure}

\begin{figure}[H]
\centering
   \includegraphics[width=1\textwidth]{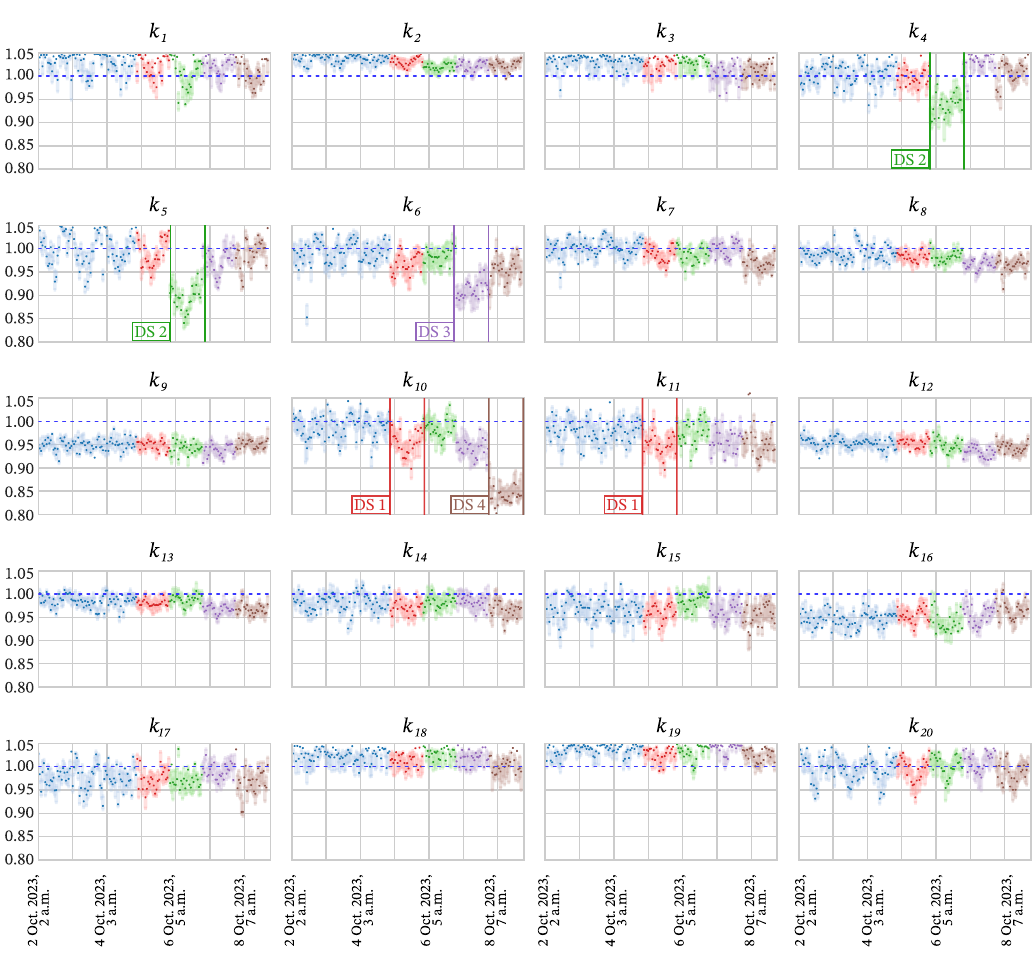}
   \caption{Median and interquartile range (25th–75th percentiles) of the posterior distributions during the period from October 2 to October 9, 2023, under the imposed damage scenarios accounting for frequency and mode shapes predictions in the likelihood. Blue represent the undamaged configuration; red corresponds to Damage Scenario 1, green to Scenario 2, violet to Scenario 3, and brown to Scenario 4.}    
    \label{fig:Posterior_means_std_freq}
\end{figure}

The variations in both natural frequencies and modal displacements were induced on a set of modes monitored on the M\'endez-N\'u\~nez Bridge between October 2 and October 9, 2023. Specifically, no damage was introduced during the first four days (October 2–5). Subsequently, \textit{Damage Scenario 1} was applied between October 5 and 6, \textit{Damage Scenario 2} between October 6 and 7, \textit{Damage Scenario 3} between October 7 and 8, and \textit{Damage Scenario 4} between October 8 and 9. The damage scenarios were applied to the recordings by scaling the frequencies according to the frequency decays reported in Table \ref{tab:frequency_decays}. Similarly, the modal displacements of each mode were scaled based on the corresponding percentage variations with respect to the undamaged condition, as assessed through the FEM. Each damage scenario analyzed in the study represents a single, isolated case of damage. They were not combined in order to clearly attribute the effects to each specific case.  A total of 24 recordings per day (one per hour) were considered, resulting in 168 samples overall. The tracked frequencies were considered subject to daily variations. No statistical model was applied to filter out environmental effects from the frequency data, as the aim was to assess the model’s ability to account for such variations and determine whether potential damage remains identifiable despite these confounding factors.

The transferred surrogate model $\hat{M}_t$ was then employed as the core of the surrogate model-based Bayesian damage identification framework. Accordingly, MCMC sampling was used to estimate the posterior distributions of the stiffness multipliers $k$, leveraging the modal data from the current observation. The initial proposal was set to 1 for each stiffness multiplier, corresponding to the undamaged condition. The prior distribution for the $i$-th damage multiplier was defined as a truncated Gaussian centered at 1,  representing the undamaged assumption $k_i = 1$, with truncation limits at $a_i=0.80$ and $b_i=1.05$ for $i = 1, \dots, N$. The likelihood was defined based on the predictions of $\hat{M}_t$, as described in Eq.~(\ref{likelihood_total}). The contribution of the mode shapes to the likelihood was computed using the MAC, as per Eq.~(\ref{likelihood_MAC}), constrained between 0 and 1 and centered around a mean value of 1, consistent with the MAC definition. Since the prior distribution was defined in terms of stiffness multipliers while the input parameters of $\hat{M}_t$ are the non-dimensional damage parameters $\pi$, each proposed sample was transformed into the $\pi$ domain before being evaluated by $\hat{M}_t$.
 
Since the damage-induced variations in terms of natural frequencies are typically more significant compared to those in terms of mode shapes, a first trial was conducted to evaluate the performance of the Bayesian-based approach when relying solely on natural frequencies for likelihood definition, thereby excluding the use of mode shape predictions in the likelihood evaluation. Here, the standard deviations for all frequencies were uniformly set to 2\%, and the prior standard deviations was fixed at 4\% for all stiffness multipliers. The proposal covariance matrix was initialized as a diagonal matrix with entries equal to $0.03\%$, scaled by the factor $s_N$ defined in Section~\ref{section22}. The total number of MCMC samples $N_s$ was set to 5000, with a burn-in phase of $N_b = 2000$.
The mode and the 25th and 75th percentile values of the posterior distributions of the stiffness multipliers during the previously defined tracking period are presented in Figure~\ref{fig:Posterior_means_std_freq}. It can be observed that fluctuations in the stiffness multipliers caused by daily temperature variations are clearly identifiable. Additionally, the damage scenarios produce distinct shifts in the stiffness patterns that significantly deviate from those observed under undamaged conditions. Therefore, the damage-induced shifts are clearly distinguishable from those caused by temperature effects. Nonetheless, although the damage-related shifts in the stiffness multipliers correctly correspond to the control regions where damage was induced, similar shifts tend to emerge in symmetric regions of the bridge. This is due to the global and symmetric nature of both the mode shapes and the structure itself, which results in increased uncertainty in damage localization. Consequently, this highlights that relying on frequency information alone is insufficient for accurate damage localization. The inclusion of mode shape information is essential to reduce spatial ambiguities and enhance the inference’s sensitivity to localized damage patterns.

Based on the findings above, in the second stage, the mode shapes are included in the likelihood evaluation. In this case, the standard deviation of the prior was set to 4\% for all stiffness multipliers, except for $k_2$, $k_9$, $k_{11}$, and $k_{19}$, for which it was reduced to 2\% to mitigate their pronounced influence on the global response and to prevent posterior distributions from drifting toward the extremes of the defined boundaries. The standard deviations for the frequencies and MACs were assigned based on a Sobol sensitivity analysis that quantified the influence of each stiffness multiplier on individual modal parameters. According to this analysis, the standard deviation for frequencies was set to 1.5\% for modes 1 to 6 and 2\% for Modes 7 and 8. For the MAC, the standard deviation was defined as 0.6\% for Modes 1, 2, 6, 7, and 8, and 1.2\% for modes 3, 4, and 5.

The results of the continuous SM-based Bayesian inference under this assumption are shown in Figure~\ref{fig:Posterior_means_std}.  All induced damages are effectively represented, with each scenario producing distinct shifts in the pattern of the corresponding stiffness multipliers that deviate noticeably from the undamaged baseline.
\textit{Damage Scenario 1}, involving a distributed stiffness reduction in the central span, resulted in modest changes in the posterior distributions. This behavior contrasts with \textit{Damage Scenario 2}, where an identical stiffness reduction was applied in a non-symmetric region, yielding more pronounced posterior deviations. This is consistent with the greater detectability of non-symmetric damages through global mode shapes. \textit{Damage Scenarios 3 and 4} produced notable shifts in the multipliers $k_6$ and $k_{10}$, respectively. Although some cross-sensitivity effects were observed (manifested as changes in non-damaged regions) larger deviations consistently occurred in the correct stiffness multipliers. This highlights the model’s ability to detect localized damage and underscores its sensitivity to both the intensity and spatial extent of the damage.

The mean computational time required to perform the surrogate-based Bayesian inference for a single time instance $t$ was approximately 3 minutes and 35 seconds. The computation was executed on a workstation equipped with an \textit{AMD Ryzen}\texttrademark{} \textit{5 7600X 6-Core} Processor operating at 4.70 GHz and 128 GB of RAM.
Overall, these results demonstrate the capability of the surrogate model $\hat{M}_t$ to serve as an efficient and scalable tool for real-time SHM and probabilistic damage assessment in large-scale bridge structures.

%%%%%%%%%%%%%%%%%%%%%%%%%%%%%%%%%%%%%%%%%%%%%

\section{Conclusions}\label{section5}

This work has presented a novel methodology for network-scale bridge Structural Health Monitoring based on a surrogate-modeling approach empowered by transfer learning.
The key contributions of this work are included in the following bullets:

\begin{itemize}
    \item A neural network-based surrogate model has been proposed to approximate the dynamic response of complex bridge structures with high predictive accuracy and computational efficiency, enabling the execution of computationally intensive Bayesian MCMC damage identification analyses within time frames compatible with continuous SHM. In the presented real-world case study, the use of FNNs reduced the computation time for Bayesian identification to only 3 minutes and 35 seconds. Such a level of efficiency is crucial for decision-making, as Bayesian approaches provide not only the probability of damage in specific structural regions but also the full distribution of damage parameters, supporting more informed and robust maintenance strategies.
    \item The model’s transferability has been initially tested on a theoretical five-span beam to verify its ability to generalize structural behavior across configurations with consistent physical properties, thus confirming the effectiveness of transfer learning in reducing the computational effort for training surrogates on similar domains.
    \item The approach has then been applied to two real-world bridges: the Volumni Bridge (\textit{source}), in Italy, and the M\'endez-N\'u\~nez Bridge (\textit{target}), in Spain. The surrogate model for the \textit{source bridge} was trained using a dataset of 1024 Monte Carlo simulations of a FEM of the bridge. Subsequently, the surrogate model of the \textit{target bridge} was successfully built via fine-tuning transfer learning from the source model, using a reduced dataset of only 256 FEM evaluations. Both surrogate models achieved coefficients of determination close to 100\% for frequencies and MAC values exceeding 0.98. This excellent predictive performance validates the effectiveness of the method in real bridge applications, proving that the proposed methodology enables the creation of highly accurate surrogate models even when direct training data for the target structure are limited, by leveraging the prior knowledge embedded in the neural network.
    \item Finally, the transferred surrogate model of the  M\'endez-N\'u\~nez Bridge has been integrated into a Bayesian-based framework for continuous damage assessment. This has enabled the identification and probabilistic quantification of four synthetically induced damage scenarios under realistic operational conditions characterized by daily temperature variations. Modal features have been directly extracted from the monitoring data without pre-processing for the isolation of thermal effects, allowing for the evaluation of the model’s robustness in practical applications and its capability to distinguish between frequency pattern variations due to thermal effects and those due to evolving damages.
    \item The proposed approach has exhibited high sensitivity to both the location and intensity of the damage. Furthermore, the results have highlighted the framework’s capability to detect different types of structural degradation, as a function of the spatial distribution of the damage, the global nature of the available mode shapes, and the influence of the bridge’s physical configuration and boundary conditions. This work has also highlighted the importance of incorporating mode shape information for accurate damage localization, as frequency-only analyses may be insufficient for reliable damage assessment.
\end{itemize}

Overall, the methodology lays the groundwork for a new paradigm in Structural Health Monitoring: that is scalable, model-driven, and informed by transferable knowledge between structurally similar assets. This opens the door to network-scale applications, where transfer learning facilitates the rapid and effective deployment of surrogate models across multiple bridges, enabling cost-effective, high-fidelity monitoring and damage assessment within increasingly data-rich infrastructure systems. This opens the door to network-scale applications, where transfer learning facilitates the rapid and effective deployment of surrogate models across multiple bridges, enabling cost-effective, high-fidelity monitoring and damage assessment within increasingly data-rich infrastructure systems. Future developments will focus on extending the proposed approach to larger networks involving more than two bridges and greater typological dissimilarities, thereby providing more generalizable conclusions and promoting broader knowledge sharing. In this context, a current limitation of the proposed methodology concerns the fixed size of the input layer, which requires the same number and type of selected damage-sensitive parameters across all structures. To address this limitation, future work will explore more flexible FNN architectures capable of handling variable input configurations, such as input-layer masking strategies or graph neural networks. Additionally, expanding the portfolio of pre-trained surrogate models will facilitate knowledge transfer across a wider range of structural typologies. A further enhancement will involve the integration of static monitoring data, such as measurements from inclinometers, to improve the accuracy of damage assessment in scenarios involving localized or subtle damage.

\section*{Acknowledgments}\label{Sec:Acknowledgeoffunding}
The authors gratefully acknowledge ANAS S.p.A. for providing access to monitoring data and technical documentation related to the Volumni Bridge. This study was supported by FABRE  - “Research consortium for the evaluation and monitoring of bridges, viaducts and other structures” (\href{www.consorziofabre.it/en}{www.consorziofabre.it/en}) within the activities of the FABRE-ANAS 2021-2026 research program. E.G.M gratefully acknowledge funding from the Spanish Ministry of Science and Innovation, the Spanish State Research Agency, and NextGenerationEU through the research project “SMARTBRIDGES-Monitorización Inteligente del Estado Estructural de Puentes Ferroviarios” (Ref. PLEC2021-007798). Any opinions expressed in this paper do not necessarily reflect the views of the funders.

\appendix
\renewcommand{\thesection}{Appendix \Alph{section}}

\bibliographystyle{elsarticle-num}

\end{document}